\documentclass[11pt,a4paper]{article}
\usepackage{times,latexsym}
\usepackage{url}
\usepackage[T1]{fontenc}
\usepackage{todonotes}
\usepackage{amsmath,amssymb}
\usepackage{booktabs}
\usepackage{graphicx}
\usepackage{subcaption}
\usepackage{enumitem}
\usepackage{float}
\usepackage{multirow}
\usepackage{xcolor}
\usepackage[most]{tcolorbox}
\usepackage{tikz}
\usetikzlibrary{arrows.meta, positioning, shapes.geometric}

\newcommand{\conceptor}{\mathbf{C}}
\usepackage[acceptedWithA]{tacl2021v1}

\usepackage{xspace,mfirstuc,tabulary}

\newif\iftaclinstructions
\taclinstructionsfalse %
\iftaclinstructions
\renewcommand{\confidential}{}
\renewcommand{\anonsubtext}{(No author info supplied here, for consistency with
TACL-submission anonymization requirements)}
\newcommand{\instr}
\fi

\iftaclpubformat %

\else

\fi

\title{Conceptors for Semantic Steering}

\author{
  \textbf{Ilias Triantafyllopoulos}$^{1}$\Thanks{Correspondence to: \texttt{ilias.triantafyllopoulos@nyu.edu}, \texttt{jsedoc@stern.nyu.edu}.}
  \quad \textbf{Young-Min Cho}$^{2}$
  \quad \textbf{Ren Tao}$^{2}$
  \quad \textbf{Miranda Muqing Miao}$^{2}$
  \\
  \textbf{Sunny Rai}$^{2}$
  \quad \textbf{Lyle Ungar}$^{2}$
  \quad \textbf{Sharath Chandra Guntuku}$^{2}$
  \quad \textbf{Neville Ryant}$^{2}$
  \quad \textbf{Jo\~ao Sedoc}$^{1}$
  \\
  \ \\
  $^{1}$New York University \qquad $^{2}$University of Pennsylvania
}

\date{}

\begin{document}
\maketitle
\begin{abstract}

Activation-based steering provides control of LLM behavior at inference time, but the dominant paradigm reduces each concept to a \emph{single direction} whose geometry is left largely unexamined. Rather than selecting a single steering direction, we use \emph{conceptors}: soft projection matrices estimated from activations pooled across both poles of a bipolar concept, which preserve the concept’s full multidimensional subspace.
A geometric analysis shows the bipolar subspace strictly subsumes the single-vector baseline. We further show that the \emph{conceptor quota} provides a parameter-free layer-selection diagnostic, predicting concept separability with Pearson correlations up to $r=0.96$ across three instruction-tuned models and three semantic dimensions. Beyond selection, conceptors admit a closed-form Boolean algebra (AND, OR, NOT): we evaluate conceptor compositionality on thematically related sub-concepts. Across a systematic five-axis design-space evaluation, conceptors match or outperform additive baselines at layers where concept subspaces are multi-dimensional while producing substantially fewer degenerate outputs. Conceptor steering is a geometrically principled, compositional, and practically safer alternative to single-direction steering from a limited number of contrastive pairs.

\end{abstract}

\section{Introduction}

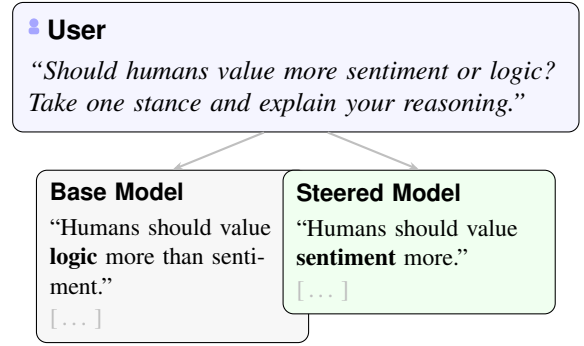
\begin{figure}[t]
\centering
\begin{tikzpicture}[
    promptbox/.style={draw, rounded corners=4pt, fill=blue!4, text width=\columnwidth-18pt, align=left, inner sep=6pt, font=\small},
    respbox/.style={draw, rounded corners=4pt, text width=0.44\columnwidth-4pt, align=left, inner sep=5pt, font=\footnotesize},
    arr/.style={-{Stealth[length=4pt]}, thick, gray!50}
]
\node[promptbox] (prompt) {%
    \begin{tikzpicture}[baseline=-0.5ex]
      \fill[blue!35] (0,0.13) circle (0.06);
      \fill[blue!35, rounded corners=1pt] (-0.07,-0.02) rectangle (0.07,0.07);
    \end{tikzpicture}\hspace{3pt}%
    \textsf{\textbf{User}}\\[3pt]
    \emph{``Should humans value more sentiment or logic? Take one stance and explain your reasoning.''}%
};

\node[respbox, fill=gray!6, below left=14pt and -5pt of prompt.south, anchor=north east] (base) {%
    \textsf{\textbf{Base Model}}\\[2pt]
    ``Humans should value \textbf{logic} more than sentiment.''\\[1pt]
    \textcolor{gray!50}{[\,\ldots\,]}%
};

\node[respbox, fill=green!6, below right=14pt and -5pt of prompt.south, anchor=north west] (steered) {%
    \textsf{\textbf{Steered Model}}\\[2pt]
    ``Humans should value \textbf{sentiment} more.''\\[1pt]
    \textcolor{gray!50}{[\,\ldots\,]}%
};

\draw[arr] ([xshift=-12pt]prompt.south) -- (base.north);
\draw[arr] ([xshift=12pt]prompt.south) -- (steered.north);
\end{tikzpicture}
\caption{Our sentiment-trained conceptor steers Gemma-2-2B-IT toward a sentiment-oriented stance while preserving fluency and instruction following.}
\label{fig:qualitative-example}
\end{figure}

Large language models encode semantic concepts as structured linear representations in their hidden states~\citep{park2024linear, gurnee2024language, sofroniew2026emotion}.
This has motivated \emph{activation-based steering} methods that modify hidden states at inference time to control model behavior without retraining: an approach that is lightweight, reversible, and attractive for safety-critical applications where controllability must be both precise and auditable.
Figure~\ref{fig:qualitative-example} illustrates this idea with a representative example: a single conceptor-based intervention shifts the model's response toward the target concept (sentiment) while preserving fluency and instruction-following.

The dominant paradigm represents each concept as a \emph{single direction}: Activation Addition~\citep{turner2023actadd} injects the mean of target-class activations, Contrastive Activation Addition~\citep{panickssery2024caa} uses the mean difference between contrastive pairs, and Inference-Time Intervention~\citep{li2023iti} selects attention heads via probing.
These methods are effective but fundamentally limited: when a concept occupies a multi-dimensional subspace, a single vector cannot capture its full geometry, and the residual structure is lost.
Moreover, selecting \emph{where} to intervene typically requires training probing classifiers at every layer, a costly supervised process that scales with model depth and concept count.

We apply \emph{conceptor matrices}~\citep{jaeger2014controlling}, soft projection operators derived from the sample correlation matrix of concept-associated activations, to semantic steering of LLM generations, addressing both limitations.
Semantic concepts such as sentiment or political leaning manifest through diverse lexical, syntactic, and discourse patterns; their neural representations therefore span multi-dimensional subspaces rather than single directions.
Rather than collapsing this structure onto one vector, a conceptor retains the full subspace, continuously attenuating each principal direction according to its signal energy (Eq.~\ref{eq:conceptor-eigen}).
Additionally, its Boolean algebra (NOT, AND, OR) enables principled composition of multiple concepts without access to original activation data.

Our contributions are as follows:
\begin{enumerate}[leftmargin=*,itemsep=0pt,topsep=0pt]
    \item \emph{Geometric characterization of concept subspaces.} We characterize key geometric properties of concept subspaces in LLMs through a unified diagnostic framework that addresses three questions: how concepts are represented, where they are expressed in the network, and how they interact with other concepts. Our findings are: 
    (i) The bipolar subspace \emph{subsumes} unipolar steering: it captures ${\geq}95\%$ of the mean-difference direction used by prior methods. (ii) The \emph{conceptor quota} is a parameter-free layer-selection diagnostic that predicts concept location at Pearson $r$ up to $0.96$. (iii) Cross-concept subspace overlap follows a universal hierarchy across architectures ($0.18$--$0.38$); we quantify how subspace overlap predicts the behavior of Boolean operations (NOT, OR) and when it limits them (AND with distant conceptors).

    \item \emph{Boolean compositionality with explicit overlap requirements.} While conceptors admit a full Boolean algebra (AND, OR, NOT), we evaluate AND and selective suppression via AND-NOT, showing that compositional steering is governed by subspace overlap: AND is more meaningful when concept subspaces share structure, while NOT (via AND-NOT) remains effective under low overlap. A pair of near-orthogonal concepts (political leaning $\times$ sentiment, overlap ${\approx}0.19$) admits a diagnostic NOT; thematically related sub-concepts (abortion $\times$ LGBTQ rights, overlap ${\approx}0.50$) yield both an informative AND \emph{and} AND-NOT.

    \item \emph{Robustness and hyperparameter sensitivity of activation steering.} Steering papers rarely report the sensitivity of their methods to practical design choices. We formalize five such choices: placement, token scope, combination rule, training source, and autoregressive mode. We systematically evaluate conceptor steering across them on three semantic concepts. Conceptors match or outperform additive baselines at layers where concept subspaces are multi-dimensional, with substantially fewer degenerate outputs ($13\%$ vs.\ $58\%$).

\end{enumerate}

\section{Background}
\label{sec:background}

This section introduces the conceptor framework originally developed by \citet{jaeger2014controlling} for controlling recurrent neural networks and describes how its core mathematical machinery can be repurposed for steering the internal representations of transformer-based language models.

\noindent
\textbf{Conceptor matrices.}
Conceptors were introduced as a mechanism for managing and selectively recalling dynamical patterns stored in recurrent neural networks. Given a collection of $N$ neural activation vectors $\{x_1, \ldots, x_N\} \subset \mathbb{R}^d$ associated with a particular concept, the conceptor matrix $\conceptor \in \mathbb{R}^{d \times d}$ is defined as the solution to the regularized reconstruction problem:
\begin{equation*}
\label{eq:conceptor-obj}
\conceptor = \arg\min_{C} \; \frac{1}{N}\sum_{i=1}^{N}\|x_i - Cx_i\|^2 + \alpha^{-2}\|C\|_F^2,
\end{equation*}
where $\|\cdot\|_F$ denotes the Frobenius norm and $\alpha > 0$ is the \emph{aperture} parameter controlling the trade-off between faithfully preserving the activation patterns and regularization. Defining the sample correlation matrix $R = \frac{1}{N} X^\top X$, where $X \in \mathbb{R}^{N \times d}$ has activation vectors as rows, the optimization admits the closed-form solution:
\begin{equation*}
\label{eq:conceptor-closedform}
\conceptor = R\left(R + \alpha^{-2}I\right)^{-1}.
\end{equation*}
This reveals the conceptor as a \emph{soft} projection operator: rather than fully retaining or discarding each direction, $\conceptor$ continuously attenuates directions according to the signal energy in $R$.

\noindent
\textbf{Aperture and eigenvalue interpretation.}
In the eigenbasis of $R = U\,\mathrm{diag}(\sigma_1,\ldots,\sigma_d)\,U^\top$, the conceptor is diagonal with entries:
\begin{equation}
\label{eq:conceptor-eigen}
\gamma_j = \frac{\sigma_j}{\sigma_j + \alpha^{-2}}, \qquad j = 1, \ldots, d,
\end{equation}
where each $\gamma_j \in [0,1]$ is a soft gating coefficient for the $j$-th principal direction. Directions with high signal energy ($\sigma_j \gg \alpha^{-2}$) are preserved ($\gamma_j \to 1$), while low-energy ones are suppressed ($\gamma_j \to 0$). The aperture $\alpha$ acts as a resolution control: large $\alpha$ causes the conceptor to approach the identity, while small $\alpha$ produces a narrow subspace projection. This continuous interpolation distinguishes conceptors from hard subspace methods such as PCA truncation or binary masks.

\noindent
\textbf{Spectral diagnostics for layer selection.}
Selecting the intervention layer is a practical challenge for any activation-based steering method. The eigenvalue structure of
$R$ is a natural starting point: a rapidly saturating spectrum indicates a low-dimensional concept subspace, a slowly decaying one indicates distributed structure. The standard summary, cumulative explained variance ratio $\mathrm{EVR}(k) = \tfrac{\sum_{j=1}^{k} \sigma_j}{\sum_{j=1}^{d} \sigma_j}$, sees only the top of the spectrum and discards medium-energy structure that we show is informative for steering.  
\citet{jaeger2014controlling} introduced the  \emph{conceptor quota} $q(\conceptor) = \tfrac{1}{d}\operatorname{tr}(\conceptor) = \tfrac{1}{d}\sum_j \gamma_j$ as a soft effective dimensionality of $\conceptor$: the average of the gating coefficients $\gamma_j$ in Eq.~\ref{eq:conceptor-eigen}, lying in $[0,1]$.
Because the gating is sigmoidal, the quota integrates the \emph{entire} spectrum, capturing intermediate-energy directions an EVR head-truncation discards.
We adopt $q(\conceptor)$ as our default diagnostic; Section~\ref{sec:finding-quota} compares it empirically to EVR.

\noindent
\textbf{Boolean operations and compositionality.}
A key advantage of the conceptor framework is the existence of principled algebraic operations that mirror Boolean logic \citep{jaeger2014controlling}. The \emph{NOT} operation, $\neg \conceptor = I - \conceptor$, suppresses a concept by passing its complementary subspace. The \emph{AND} operation, $\conceptor_{A} \wedge \conceptor_B = (\conceptor_A^{-1} + \conceptor_B^{-1} - I)^{-1}$, yields a conceptor retaining only directions supported by both operands---effectively their subspace intersection---while the \emph{OR} operation, defined via De Morgan's law, captures the union of two concept subspaces. Crucially, all three operations require only the conceptor matrices themselves, with no access to the original activation data,
making them well-suited for combining conceptors across different corpora. While this algebra is complete, our empirical evaluation focuses on AND and NOT (via AND-NOT), which most directly expose the role of subspace overlap; OR behavior follows from the same geometry via De Morgan’s law.

\noindent
\textbf{Application to transformers.}
Conceptors were originally developed for echo state networks \citep{jaeger2001echo}, where they gated temporal patterns in a fixed recurrent reservoir. We adapt them to transformers~\citep{vaswani2017attention} by using concept-associated hidden states in place of reservoir trajectories and applying the conceptor as a forward-pass hook at a chosen layer during generation. This relies on the shared structure of both systems: high-dimensional distributed activations in which concepts occupy overlapping but distinguishable subspaces, while leaving the mathematical core intact.

\noindent
\textbf{Bipolar versus unipolar steering.}
To capture the full extent of a concept's subspace, we train each conceptor on the \emph{union of both poles}, pooling positive \emph{and} negative sentiment activations into a single correlation matrix.
This is a key departure from prior work: additive methods compute a difference vector between poles~\citep{turner2023actadd, panickssery2024caa} or steer toward a single target pole, and prior conceptor work trains separate operators per direction~\citep{postmus2024steering}, all inherently unipolar. Conceptors themselves are agnostic to this choice, but for inherently bipolar concepts such as sentiment or political leaning, pooling both poles captures the full between-pole subspace that single-pole training (e.g., a ``positive'' rather than a general ``sentiment'' conceptor) misses.

\section{Related Work}
\label{sec:related}

\subsection{Activation-Based Steering}

Existing methods for steering language models via internal representations span a spectrum from simple additive interventions to learned subspace edits, but all require either supervised training or collapse concepts onto a single direction.
Activation Addition \citep{turner2023actadd} and Contrastive Activation Addition \citep{panickssery2024caa} inject a single mean-difference vector into the residual stream; Representation Surgery \citep{singh2024repsurgery} generalizes these by optimally transforming activations to match both the mean and covariance of concept distributions. Inference-Time Intervention \citep{li2023iti} and CORAL \citep{miao2026coral} train probes on hidden activations to identify and shift informative directions. Recent geometric methods such as Spherical Steering \citep{you2026spherical} argue against magnitude distortion and steer via norm-preserving geodesic rotations.
Concurrently, \citet{sofroniew2026emotion} demonstrate that emotion concepts in Claude Sonnet~4.5 are represented as linear directions recoverable via mean-difference vectors, and that steering along these directions causally shifts model behavior in forced-choice evaluations.
Their work uses 171 unipolar emotion vectors; by contrast, our bipolar conceptor formulation captures the full concept subspace in a single operator, subsuming individual polar directions.
Conceptors unify these geometric intuitions with second-order statistical constraints; they use regularized correlation structures to quiet noisy dimensions without squashing the manifold. This framework characterizes multi-dimensional concept volumes rather than single polar axes, enabling more expressive and algebraically compositional control.
Unlike the polar or asymmetric mappings found in nearly every steering method, conceptors function as bipolar filters that support training-free, algebraically compositional steering via principled Boolean logic.
More expressive approaches such as LoReFT \citep{wu2024reft} learn low-rank interventions on frozen representations via gradient-based optimization.
Sparse autoencoders \citep{bricken2023monosemanticity, templeton2024scaling} decompose activations into interpretable features that can be individually amplified, but require substantial training overhead and lack algebraic compositionality.

\subsection{Conceptor Theory and Prior Applications}
\label{sec:related-conceptors}

Conceptors were introduced by \citet{jaeger2014controlling} for managing dynamical patterns in echo state networks \citep{jaeger2001echo} and later formalized by \citet{jaeger2017using}.
In NLP, conceptor negation has been used to remove uninformative variance from word embeddings \citep{liu2019unsupervised} and to debias representations \citep{karve2019conceptor}, while the Boolean OR operation has supported continual learning of sentence representations \citep{liu2019continual}; \citet{he2018overcoming} extended the framework to deep feedforward networks.
Most directly related to our work, \citet{postmus2024steering} applied conceptors as steering matrices for GPT-J and GPT-NeoX, showing improvements over additive baselines and effective multi-concept composition via Boolean AND.
We substantially extend this line of research with a parameter-free spectral diagnostic for layer selection, a systematic five-factor design-space exploration, and evaluation across broader semantic dimensions.

\subsection{Layer Selection and Evaluation}

Selecting where in a network to intervene is typically done by training probing classifiers at every layer and choosing the best \citep{belinkov2022probing}, a process whose cost scales with model depth and concept count.
Prior probing studies have established that semantic properties peak in middle layers \citep{tenney2019bert, conneau2018you}, a pattern our spectral diagnostic recovers without any supervised infrastructure.

\section{Spectral Geometry of Conceptor Representations}
\label{sec:spectral-analysis}

We analyze the spectral structure of conceptor representations to provide geometric evidence for their utility as steering operators.
Our analysis spans three instruction-tuned models---Gemma-2-2B-IT~\citep{gemma2}, Gemma-2-9B-IT~\citep{gemma2}, and Qwen-2.5-3B-Instruct~\citep{qwen2.5}---and three semantic concepts (sentiment, political leaning, depression).
For each model--concept--layer triple, we extract post-block hidden states from 100 contrastive sentence pairs (200 sentences total), compute the correlation matrix, $R$, and derive conceptor matrix, $\conceptor$.

\subsection{Conceptor Quota Predicts Concept Separability}
\label{sec:finding-quota}

Selecting intervention layers in transformers typically requires training probing classifiers at every layer~\citep{belinkov2022probing}, a costly supervised procedure. We show that the \emph{conceptor quota}, computed directly from the steering operator, provides a parameter-free alternative.

Recall the conceptor quota $q(\conceptor)=\tfrac{1}{d}\operatorname{tr}(\conceptor)$ defined in Section~\ref{sec:background}: a soft effective dimensionality of $\conceptor$ that integrates the full spectrum of the activation correlation matrix $R$ through the sigmoidal gating of Eq.~\ref{eq:conceptor-eigen}.

Table~\ref{tab:quota-vs-evr} compares quota and EVR by correlating each with probe AUC from $\ell_2$-regularized logistic regression classifiers trained and evaluated on external data: SST-2, Rotten Tomatoes, and TweetEval for sentiment; Hyperpartisan News for political leaning; and Reddit Depression and Clinical Depression for depression (Appendix~\ref{sec:appendix-probe-datasets}). On Gemma-2-2B-IT, the two diagnostics are comparably effective. On Gemma-2-9B-IT, quota correlates more strongly with AUC than EVR. On Qwen, where EVR is largely non-diagnostic ($r\leq0.30$) because activations collapse into very few principal components, quota reaches $r{=}0.84$--$0.96$ on all benchmarks except Clinical Depression, outperforming EVR by 0.824 on average.

The depression benchmarks clarify when such correlations are meaningful. On Reddit Depression, probe AUC is near-ceiling across layers (AUC $\approx0.96$--$0.99$), so both diagnostics correlate well with the remaining structured layer-wise variation. On Clinical Depression, probe AUC is near-chance and nearly flat (AUC $\approx0.60$--$0.64$), leaving too little variation for either diagnostic to predict. Thus, quota is informative when probe performance varies meaningfully across layers, but no spectral diagnostic can recover separability when the concept is too diffuse or the dataset too noisy for probes to separate at any layer.

This result is robust across apertures $\alpha\in[2,50]$ (Appendix, Figure~\ref{fig:alpha-heatmap}). The quota therefore provides a practical unsupervised layer-selection tool requiring only ${\sim}100$ contrastive pairs, one eigendecomposition per layer, and modest compute.\footnote{In our experiments, under 30 seconds on one L40S GPU.}

\begin{table}[t]
\centering
\small
\begin{tabular}{ll cc}
\toprule
\textbf{Model} & \textbf{Test set} & $\boldsymbol{r_{\mathrm{EVR}}}$ & $\boldsymbol{r_{\mathrm{Quota}}}$ \\
\midrule
\multirow{6}{*}{Gemma-2B}
  & SST-2          & \textbf{0.88} & 0.86 \\
  & Rotten Tom.    & 0.82 & \textbf{0.88} \\
  & TweetEval      & \textbf{0.92} & 0.83 \\
  & Hyperpartisan  & \textbf{0.95} & 0.71 \\
  & Reddit Depression  & 0.72 & \textbf{0.85} \\
  & Clinical Depression  & \textbf{0.56} & 0.15 \\
\midrule
\multirow{6}{*}{Gemma-9B}
  & SST-2          & 0.71 & \textbf{0.91} \\
  & Rotten Tom.    & 0.72 & \textbf{0.90} \\
  & TweetEval      & 0.63 & \textbf{0.94} \\
  & Hyperpartisan  & 0.61 & \textbf{0.95} \\
  & Reddit Depression  & 0.50 & \textbf{0.93} \\
  & Clinical Depression  & 0.49 & \textbf{0.77} \\
\midrule
\multirow{6}{*}{Qwen-3B}
  & SST-2          & 0.08 & \textbf{0.84} \\
  & Rotten Tom.    & 0.04 & \textbf{0.89} \\
  & TweetEval      & 0.17 & \textbf{0.90} \\
  & Hyperpartisan  & 0.08 & \textbf{0.89} \\
  & Reddit Depression  & $-$0.01 & \textbf{0.96} \\
  & Clinical Depression  & \textbf{0.30} & 0.07 \\
\bottomrule
\end{tabular}
\caption{Pearson correlation between layer-wise spectral diagnostic and probe AUC.
$r_{\mathrm{EVR}}$ uses the top-10 explained variance ratio of $R$;
$r_{\mathrm{Quota}}$ uses the conceptor quota $q(\conceptor)$ at $\alpha{=}10$.
All quota correlations are significant ($p < 0.01$) except Clinical Depression; EVR correlations on Qwen are not significant.}
\label{tab:quota-vs-evr}
\end{table}

\subsection{Bipolar Subspace Geometry}
\label{sec:finding-bipolar}

A core design choice is training each conceptor on the \emph{union of both poles} of a bipolar concept.
Prior conceptor-based steering work~\citep{postmus2024steering} trained separate unipolar conceptors---one per target function (e.g., antonym, translation)---from activations of a single class of demonstrations, without any contrastive or bipolar structure.
We validate our method geometrically by measuring the capture fraction of the DiffMean steering vector $\Delta\bar{v} = \bar{v}_+ - \bar{v}_-$~\citep{panickssery2024caa} by the top-$k$ subspace of bipolar versus pole-specific activations:
\begin{equation}
\text{Capture}(\Delta\bar{v}, \mathbf{V}_k) = \frac{\| \mathbf{V}_k \mathbf{V}_k^\top \Delta\bar{v} \|^2}{\| \Delta\bar{v} \|^2},
\label{eq:capture}
\end{equation}
where $\mathbf{V}_k \in \mathbb{R}^{H \times k}$ is the matrix of top-$k$ right singular vectors of the activation matrix $\mathbf{X}$, and $\mathbf{V}_k \mathbf{V}_k^\top$ is the corresponding orthogonal projector.

\begin{figure}[t]
    \centering
    \includegraphics[width=\columnwidth]{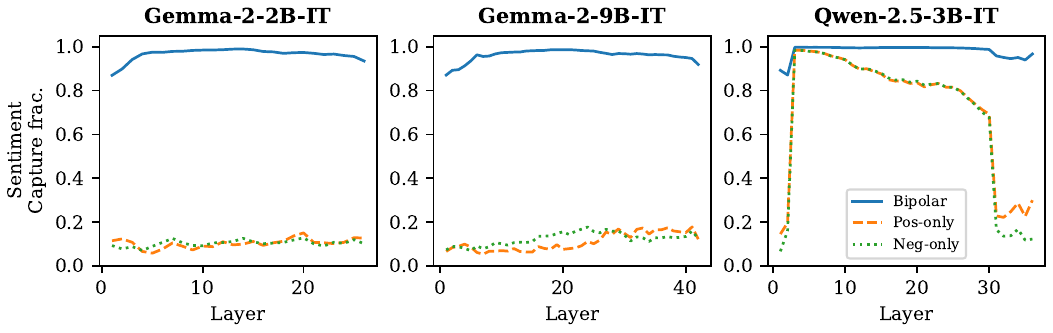}
    \caption{Fraction of the DiffMean vector captured by the top-10 subspace of bipolar (solid), pos-only (dashed), and neg-only (dotted) training, shown for sentiment across the three models.
    The bipolar subspace captures ${\geq}95\%$ on Gemma, while each pole alone captures ${\sim}10\%$. Political and depression versions are deferred to Appendix~\ref{sec:appendix-diffmean-capture-full}.}
    \label{fig:diffmean-capture}
\end{figure}

Figure~\ref{fig:diffmean-capture} reveals an order-of-magnitude gap for sentiment in Gemma models: the bipolar subspace captures ${\geq}95\%$ of $\Delta\bar{v}$, while each pole alone captures ${\sim}10\%$. The same pattern holds for depression on Gemma and is weaker but consistent on Qwen and for political leaning (full $3\times3$ grid in Appendix~\ref{sec:appendix-diffmean-capture-full}, Figure~\ref{fig:diffmean-capture-full}).
This occurs because the poles share moderate structure (overlap ${\approx}0.50$--$0.67$) and $\Delta\bar{v}$ lies in the between-pole region that only bipolar training captures.
The bipolar conceptor thus geometrically captures the bulk of single-vector methods.
This means any effect achievable by additive steering is also achievable by the conceptor, while the conceptor additionally retains stylistic, contextual, and co-occurring semantic dimensions that single-vector baselines lose.
Section~\ref{sec:steering-baselines} empirically confirms that conceptors match or exceed additive methods while maintaining higher output quality.

\subsection{Cross-Concept Independence and Concept Dimensionality}
\label{sec:finding-cross-concept}

Boolean compositionality (Section~\ref{sec:steering-compositionality}) depends on how much concept subspaces overlap. We measure cross-concept overlap as the mean squared cosine between the top-$k$ principal subspaces of two concepts~\citep{knyazev2002principal}:
\begin{equation}
\text{Overlap}(\mathbf{V}_k^{(A)}, \mathbf{V}_k^{(B)}) =
\frac{1}{k} \| {\mathbf{V}_k^{(A)}}^\top \mathbf{V}_k^{(B)} \|_F^2,
\label{eq:overlap}
\end{equation}
where $\mathbf{V}_k^{(A)}, \mathbf{V}_k^{(B)} \in \mathbb{R}^{H \times k}$ contain the top-$k$ right singular vectors of each concept's activation matrix. This is the mean squared cosine of the $k$ principal angles: 1 for identical subspaces and 0 for orthogonal ones.

\begin{figure}[t]
    \centering
    \includegraphics[width=0.45\textwidth]{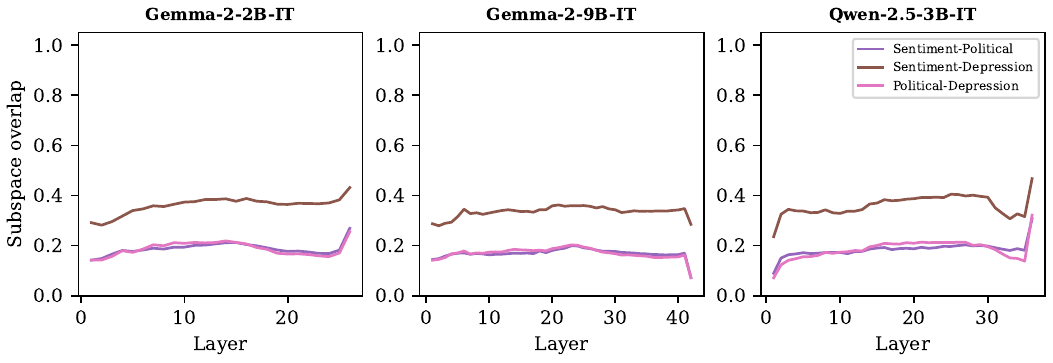}
    \caption{Cross-concept subspace overlap (top-10, mean squared cosine).
    Low values follow a universal hierarchy across architectures; they enable NOT and OR to preserve cross-concept structure while requiring thematic relatedness for AND.}
    \label{fig:cross-concept-overlap}
\end{figure}

Figure~\ref{fig:cross-concept-overlap} shows low top-10 overlap across all three architectures: Sentiment--Political ${\approx}0.19$, Political--Depression ${\approx}0.20$, and Sentiment--Depression ${\approx}0.34$. The hierarchy is linguistically expected: sentiment and depression share affective valence, whereas political leaning depends more on policy, ideology, and partisanship. Low overlap has opposite implications for different Boolean operations. It makes NOT($\conceptor_\text{sent}$) preserve ${\geq}80\%$ of political and depression structure because the subspaces barely intersect, but leaves too little shared structure for meaningful AND between distant concepts. Section~\ref{sec:steering-compositionality} quantifies this trade-off and motivates using thematically related sub-concepts for AND.

The conceptor trace $\operatorname{tr}(\conceptor)$ (unnormalized quota) at $\alpha{=}10$, averaged over layers, gives the effective number of activation directions claimed by the conceptor (out of at most $N{=}200$). The ordering sentiment $<$ political $<$ depression holds on all three architectures (Gemma-2-2B: $196.4 / 197.3 / 197.8$; Gemma-2-9B: $197.7 / 197.9 / 198.3$; Qwen-2.5-3B: $188.7 / 191.5 / 193.8$), tightly saturated on Gemma and most pronounced on Qwen, consistent with concept complexity rising from a single-axis sentiment construct to a multi-faceted clinical one.

\section{Conceptor-Based Semantic Steering}
\label{sec:contribution-steering}

We now use conceptors as inference-time steering operators. We evaluate sentiment (positive/negative valence), political leaning (left/right), and depression (depressive/healthy language) on Gemma-2-2B-IT and Gemma-2-9B-IT. We first show that conceptors outperform additive baselines on a per-layer sweep (Section~\ref{sec:steering-baselines}), then demonstrate Boolean compositionality on two concept pairs of contrasting subspace overlap (Section~\ref{sec:steering-compositionality}), examine robustness across five design choices (Section~\ref{sec:steering-guidelines}), and finally test generalization to a larger held-out set and a structured forced-choice task (Section~\ref{sec:steering-generalization}).
As discussed in Section~\ref{sec:finding-bipolar}, our bipolar conceptors are trained on the union of both poles, so they represent the full concept subspace rather than a single separating direction; geometrically, this subspace contains the DiffMean direction plus surrounding multi-dimensional structure.

\paragraph{Protocol and scoring.}
Each conceptor is trained on 100 contrastive sentence pairs from both poles. At a single intervention layer $\ell$, we steer hidden states using replacement or interpolation (see Q3):
\begin{align}
\text{Replace:}\quad z' &= \beta\, C\, z, \label{eq:replace} \\
\text{Interpolation:}\quad z' &= (1{-}\beta)\, z + \beta\, C\, z, \label{eq:interp}
\end{align}
where $z \in \mathbb{R}^H$ is the hidden state at layer $\ell$---either residual stream or attention output, depending on placement (see Q1)---$C \in \mathbb{R}^{H \times H}$ is the conceptor trained at that layer, and $\beta$ controls steering strength. We score by \emph{win ratio}\footnote{This is for the open-ended generation task. The forced-choice MCQ task introduced in Section~\ref{sec:steering-compositionality} uses category-level probabilities and choice rates instead.}: the fraction of test prompts where the steered output scores higher than the base output on the target dimension using task-specific classifiers (Appendix~\ref{sec:appendix-prompts}).\footnote{All code, data, and evaluation scripts will be released upon acceptance.}

We compare against two additive baselines:  \emph{Addition} uses
$z' = z + \beta \bar{v}$, where
$\bar{v} = \frac{1}{N}\sum_{i=1}^{N} x_i$
is the mean activation over all $N$
training sentences at layer $\ell$, shifting $z$ toward the concept centroid. \emph{DiffMean}~\citep{panickssery2024caa} uses
$z' = z + \beta(\bar{v}-\bar{v}_{\text{null}})$,
where $\bar{v}$ averages over both poles and $\bar{v}_{\text{null}}$ averages over random sentences.\footnote{The standard DiffMean formulation~\citep{panickssery2024caa} subtracts the negative-pole mean, $\bar{v}_+ - \bar{v}_-$, producing a unipolar vector that favors one pole. Since our concepts are bipolar, we subtract a concept-neutral baseline to preserve pole symmetry. For sentiment, for example, a unipolar vector would boost positive responses but suppress negative ones, whereas our goal is to steer toward the \emph{concept} rather than a specific pole (See Appendix~\ref{sec:appendix-unipolar-diffmean}).}

\subsection{Conceptors Outperform Additive Baselines}
\label{sec:steering-baselines}

We compare all three methods across model layers, selecting the best available hyperparameters per layer from the sweep grid ($\alpha,\beta$ for conceptors; $\beta$ for additive baselines; Appendix~\ref{sec:appendix-sweep}) on a 50-sample test set.

\begin{figure}[!tp]
    \centering
    \includegraphics[width=\columnwidth]{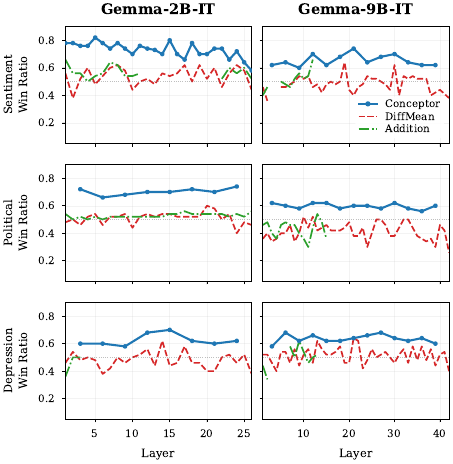}
    \caption{Best win ratio per layer for Conceptor, DiffMean, and Addition across concepts and models. Layers where all configurations produced incoherent outputs are omitted; the conceptor sweep did not evaluate every layer. Addition is degenerate in the majority of configurations (58\%).}
    \label{fig:layer-sweep}
\end{figure}

Figure~\ref{fig:layer-sweep} shows a consistent advantage for conceptors. On Gemma-2-2B-IT, they reach win ratios of 0.70--0.82 across most layers, while both additive baselines remain near 0.50--0.62. The gap is largest where the concept subspace is multi-dimensional: for sentiment, quota peaks in early-to-mid layers (3--10), matching the layers where conceptors most outperform additive baselines; for political leaning, the richer subspace appears later (15--24), and the advantage shifts accordingly. When the concept collapses to a near-one-dimensional representation, as indicated by high EVR@1, conceptors and additive methods converge because a single vector is sufficient. Qualitative examples in Appendix~\ref{sec:appendix-qualitative} show that the best sentiment conceptor can reverse the model's argumentative stance while preserving fluency.

Output stability is the main practical difference. Additive baselines often produce \emph{degenerate outputs}: repetitive and incoherent text that fills the generation budget. We classify a configuration as degenerate when its mean output length is more than twice the corresponding base-model length,
$\bar{T}_{\text{steered}}/\bar{T}_{\text{base}}>2.0$.
This captures the dominant failure mode while ignoring minor length changes from legitimate content variation, and we verify it by manual inspection. Addition is degenerate in 58\% of configurations, DiffMean in 30\%, and conceptor interpolation in only 13\%. This suggests that conceptors steer within the model's natural activation manifold, whereas additive perturbations---especially Addition, which injects an absolute mean vector rather than a contrastive direction---can push hidden states out of distribution and disrupt generation.

\subsection{Boolean Compositionality}
\label{sec:steering-compositionality}

Beyond outperforming additive baselines on individual concepts, conceptors support algebraic composition via the Boolean operations introduced in Section~\ref{sec:background}: closed-form AND, NOT, and OR on conceptor matrices, without retraining and without access to the original training activations. This is a capability single-direction methods cannot provide on a principled basis. We evaluate compositionality in two concept pairs chosen to span the cross-concept overlap regime characterized in Section~\ref{sec:finding-cross-concept}.

\begin{figure*}[!t]
    \centering
    \begin{minipage}[t]{0.48\textwidth}
        \centering
        \includegraphics[width=\textwidth]{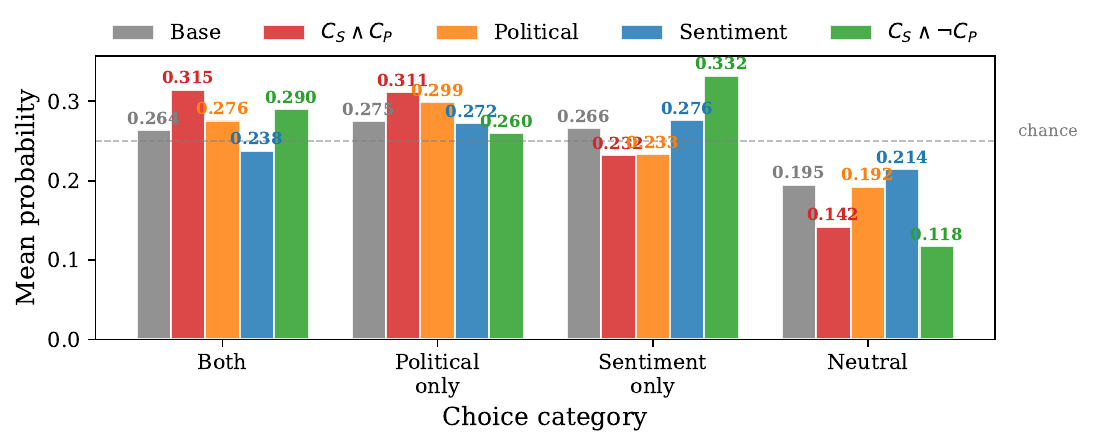}
        \\ \textbf{(a)} Political $\times$ Sentiment
    \end{minipage}
    \hfill
    \begin{minipage}[t]{0.48\textwidth}
        \centering
        \includegraphics[width=\textwidth]{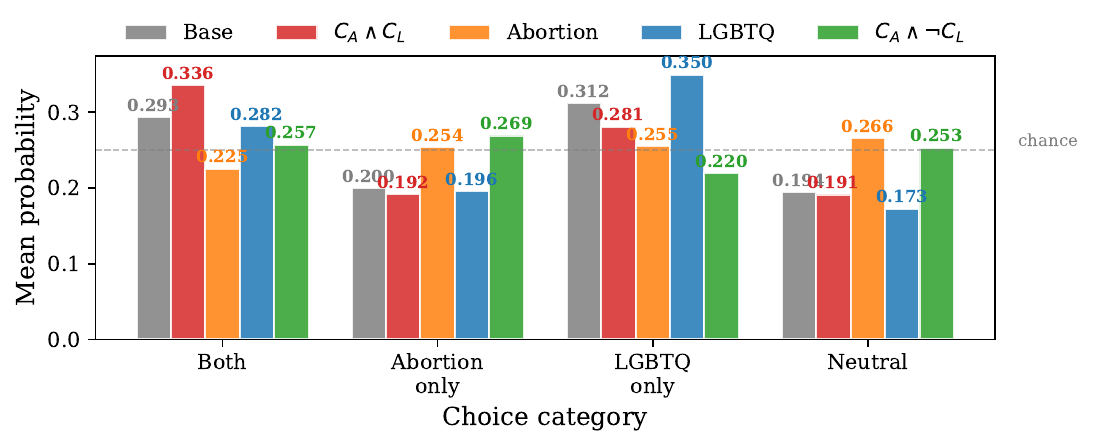}
        \\ \textbf{(b)} Abortion stance $\times$ LGBTQ rights
    \end{minipage}
    \caption{Boolean compositionality via MCQ evaluation on Gemma-2-2B-IT (best config per method): mean probability assigned to each choice category. The AND-NOT operation ($C_S \wedge \neg C_P$ in (a); $C_A \wedge \neg C_L$ in (b)) demonstrates concept-selective isolation, especially under higher subspace overlap (b). Corresponding model-final-choice rates are reported in Appendix~\ref{sec:appendix-compositionality-rates}, Figure~\ref{fig:mcq-compositionality-rates}.}
    \label{fig:mcq-compositionality}
\vspace{-0.3cm}
\end{figure*}

\paragraph{Forced-choice MCQ task.}
Open-ended generation makes multi-concept effects hard to disentangle, since each dimension requires a separate classifier and pole interactions are confounded by fluency degradation.
We therefore introduce a forced-choice multiple-choice question (MCQ) evaluation that reads behavior directly from the logit distribution over answer-letter tokens. Following recent forced-choice steering analyses~\citep{sofroniew2026emotion,panickssery2024caa} but adapting the setup to bipolar compositional selectivity, each question presents four randomly shuffled options: one aligned with \emph{both} concepts, one with each single concept, and one neutral distractor. The model's choice is read from the A/B/C/D logits after a single forward pass.
We construct 300 questions per concept pair with OpinionsQA-inspired topics~\citep{santurkar2023opinions}, using generic preference prompts such as ``what should be the federal government's top priority?'' paired with the four options.
For each pair we compare the base model, AND of the two conceptors, each single-concept conceptor, and AND-NOT (one concept while suppressing the other).

\paragraph{Overlap-driven pair selection.}
The AND of two conceptors retains only directions jointly supported by both operands, so near-orthogonal concepts produce a small active subspace by construction. Conversely, the same low overlap makes NOT and OR well-behaved, since removing one concept leaves the other largely intact. This motivates a two-pair design: political~$\times$~sentiment tests the geometric limit of AND and the selectivity of NOT under low overlap, while abortion stance~$\times$~LGBTQ rights tests whether a non-trivial intersection supports genuine multi-concept steering. Appendix~\ref{sec:appendix-cross-concept-ellipses}, Figure~\ref{fig:cross-concept-ellipses}, visualizes the resulting subspaces as 2D ellipses and shows that political--sentiment ellipses intersect in only a small central region whereas abortion--LGBTQ share a substantial interior.

\paragraph{political $\times$ sentiment.}
Political and sentiment subspaces have limited overlap (${\approx}0.19$; Section~\ref{sec:finding-cross-concept}), so $\conceptor_P \wedge \conceptor_S$ should be close to degenerate, whereas $\conceptor_S \wedge \neg \conceptor_P$ should sharpen sentiment selectivity by suppressing residual political leakage. We construct the 300 MCQ questions with balanced poles (left/right $\times$ positive/negative across all combinations) and evaluate AND$(C_P,C_S)$, political-only conceptor, sentiment-only conceptor, and AND-NOT$(C_S,\neg C_P)$ (in addition to the base model).

Figure~\ref{fig:mcq-compositionality}(a) shows the results.
AND modestly raises ``Both'' (probability 0.32 vs.\ 0.26 base, without the model choice percentage increasing, as shown in Fig.~\ref{fig:mcq-compositionality-rates}) and ``Political only'' (0.31 vs.\ 0.28), while slightly reducing ``Sentiment only'' (0.23 vs.\ 0.27). Single-concept conceptors boost their own targets, and sentiment-only is the only single-concept method to raise ``Sentiment only'' (0.28 vs.\ 0.27). AND-NOT gives the strongest selective effect: it pushes ``Sentiment only'' to 0.33---the largest single-category probability in panel~(a)---while keeping ``Political only'' near base. It also suppresses ``Neutral'' most strongly (0.12 vs.\ 0.20 base). Thus, even under low overlap, NOT sharpens selectivity by reallocating mass away from neutral distractors rather than by removing a large shared subspace.

\paragraph{abortion stance $\times$ LGBTQ rights.}
We next test a pair with shared structure. Abortion stance and LGBTQ rights are social-values concepts with overlapping political valence; their top-10 overlap at layer 15 is ${\approx}0.50$, more than double political--sentiment, allowing AND to occupy a non-degenerate region. This pair is therefore chosen to match the overlap requirement above. We evaluate AND, abortion-only conceptor, LGBTQ-only conceptor, and AND-NOT$(C_A,\neg C_L)$, which targets abortion while suppressing LGBTQ signal (in addition to the base model).

Figure~\ref{fig:mcq-compositionality}(b) shows clear concept-selective effects. AND raises the ``Both'' probability (0.34 vs.\ 0.29 base), confirming joint activation. LGBTQ-only pushes ``LGBTQ only'' to 0.35---the strongest single-category effect---while abortion-only raises its target to 0.25. AND-NOT produces the strongest targeted shift in the figure: ``Abortion only'' rises from 0.20 to 0.27, while ``LGBTQ only'' drops from 0.31 to 0.22. This shows that NOT removes the LGBTQ component from the shared social-values subspace, isolating abortion-specific signal in a way unavailable to single-concept steering.

\subsection{Robustness of Conceptor Steering}
\label{sec:steering-guidelines}

Deploying conceptor steering requires navigating a complex design space. To ensure our guidelines reflect a functional regime, we evaluate sensitivity within an effective operating envelope, filtering out extreme intensities that induce model collapse. Furthermore, a sensitivity analysis reveals that architectural placement, conceptor combination, and injection mode are the foundational design choices, dictating the stability landscape for all subsequent hyperparameter tuning. We formalize five questions and provide empirically grounded answers from a systematic evaluation summarized in Table~\ref{tab:design-space-merged}. The conclusions remain intact for Depression concept and Gemma-2-9B-IT, and a complementary feature-importance analysis  (Appendix~\ref{sec:appendix-feature-importance}, Figure~\ref{fig:feature_importance_win_ratio}) corroborates that placement, combination, and injection mode dominate.

\begin{table}[t]
\centering
\small
\setlength{\tabcolsep}{3pt}
\resizebox{\columnwidth}{!}{%
\begin{tabular}{@{}ll cc cc | c @{}}
\toprule
& & \multicolumn{2}{c}{\textbf{Sentiment}} & \multicolumn{2}{c}{\textbf{Political}} & \\
\cmidrule(lr){3-4} \cmidrule(lr){5-6}
\textbf{Design choice} & \textbf{Level} & \textbf{Best} & \textbf{Median} & \textbf{Best} & \textbf{Median} & $\boldsymbol{|r_{rb}|}$ \\
\midrule
\multirow{2}{*}{Q1: Placement}
 & Residual stream    & 0.76 & \textbf{0.50} & 0.72 & \textbf{0.52} & \multirow{2}{*}{0.18} \\
 & Attention output   & \textbf{0.82} & 0.43 & \textbf{0.74} & 0.40 & \\
\midrule
\multirow{2}{*}{Q2: Token scope}
 & Last token         & 0.80 & 0.46 & \textbf{0.74} & 0.44 & \multirow{2}{*}{0.14} \\
 & All tokens         & \textbf{0.82} & \textbf{0.48} & \textbf{0.74} & \textbf{0.48} & \\
\midrule
\multirow{2}{*}{Q3: Combination}
 & Interpolation      & 0.80 & 0.40 & 0.68 & 0.38 & \multirow{2}{*}{0.38} \\
 & Replace            & \textbf{0.82} & \textbf{0.52} & \textbf{0.74} & \textbf{0.50} & \\
\midrule
\multirow{2}{*}{Q4: Training source}
 & Mean-pooled        & \textbf{0.82} & 0.48 & 0.72 & 0.46 & \multirow{2}{*}{0.08} \\
 & Last-token         & 0.80 & 0.48 & \textbf{0.74} & 0.46 & \\
\midrule
\multirow{2}{*}{Q5: Autoregressive}
 & No (Steer once)    & \textbf{0.82} & 0.44 & 0.72 & 0.42 & \multirow{2}{*}{0.26} \\
 & Yes (Steer gen.)   & 0.80 & \textbf{0.50} & \textbf{0.74} & \textbf{0.48} & \\
\bottomrule
\end{tabular}
}
\caption{Design space (Gemma-2-2B-IT): best and median win ratio for sentiment and political leaning. Bold indicates the best level per metric. Effect size $|r_{rb}|$ across all datasets quantifies the impact of each parameter.}
\label{tab:design-space-merged}
\end{table}

\noindent
\textbf{Q1 -- Where to apply the conceptor?} Pre-block residual-stream placement is more robust across hyperparameters (median 0.50 vs. 0.43 for sentiment); post-block attention-output placement reaches higher peaks but is more sensitive to configuration.

\noindent
\textbf{Q2 -- Which tokens to steer?} All-token steering is marginally better than steering the last-token only ($+0.02$--$0.04$ median).

\noindent
\textbf{Q3 -- How to combine conceptor and activation?}
This is the most impactful factor ($p < 0.001$, $r_\mathrm{rb}{=}0.38$).
Replace (Eq.~\ref{eq:replace}) achieves substantially higher median win ratio than interpolation (Eq.~\ref{eq:interp}), but produces degenerate outputs in 38\% of configurations versus 13\%, a safety-effectiveness tradeoff.

\noindent
\textbf{Q4 -- Which activations to train on?}
Mean-pooled and last-token sources perform comparably ($p = 0.58$, n.s.); either is suitable to train the conceptor.

\noindent
\textbf{Q5 -- Steer once or persistently?}
Persistent steering (applying the conceptor at every decoding step) improved median win ratio by $+0.06$.

\paragraph{Summary.}
Based on our large-scale sweep, we recommend: 
(1) residual-stream placement for robustness, or attention-output only when targeting maximum performance; 
(2) all-token steering; 
(3) the Replace operator for maximum steering, or Interpolation to prioritize safety; 
(4) any training source of conceptors; 
(5) autoregressive steering, compute permitting.

\subsection{Generalization}
\label{sec:steering-generalization}

\subsubsection{Generalization to Larger Test Sets}
\label{sec:steering-gen-large}

\begin{figure}[t]
    \centering
    \includegraphics[width=\columnwidth]{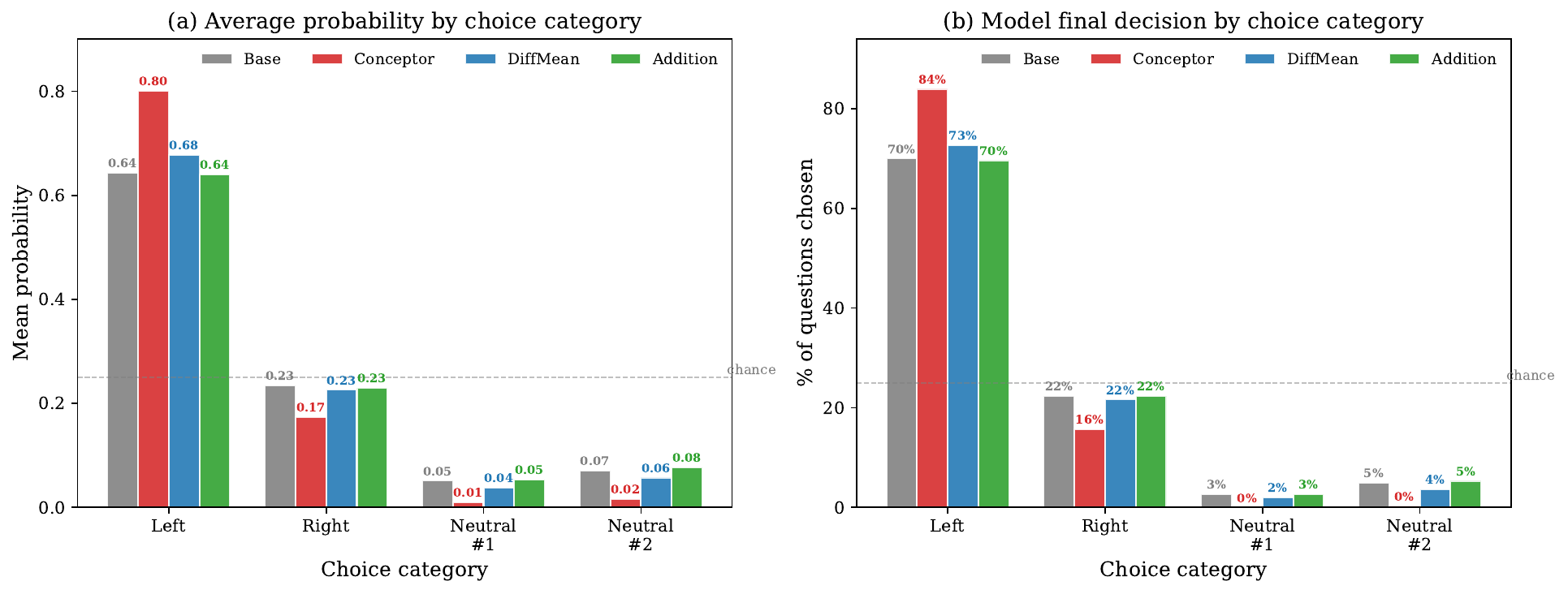}
    \caption{MCQ evaluation for political leaning on Gemma-2-2B-IT (300 questions, best config per method). (a)~Mean probability assigned to each choice category. (b)~Fraction of questions where each category was selected.}
    \label{fig:mcq-political}
\end{figure}

We examine generalization along two axes: (a)~the \emph{open-ended generation} task from Section~\ref{sec:steering-baselines}, scaled to 500 test prompts (Table~\ref{tab:generalization}), and (b)~a single-concept variant of the MCQ task introduced for compositionality in Section~\ref{sec:steering-compositionality} (Fig.~\ref{fig:mcq-political}).

\begin{table}[t]
\centering
\small
\setlength{\tabcolsep}{4pt}
\begin{tabular}{@{}l ccc ccc@{}}
\toprule
& \multicolumn{3}{c}{\textbf{Gemma-2-2B-IT}} & \multicolumn{3}{c}{\textbf{Gemma-2-9B-IT}} \\
\cmidrule(lr){2-4} \cmidrule(lr){5-7}
\textbf{Method} & \textbf{Affect} & \textbf{Pol.} & \textbf{Dep.} & \textbf{Affect} & \textbf{Pol.} & \textbf{Dep.} \\
\midrule
Conceptor & \textbf{.68} & \textbf{.53} & \textbf{.54} & \textbf{.72} & \textbf{.60} & \textbf{.57} \\
Addition  & .56 & .50 & .47 & .61 & .51 & .55 \\
DiffMean  & .56 & .50 & .50 & .56 & .48 & .55 \\
\bottomrule
\end{tabular}
\caption{Win ratio on 500-sample held-out test sets. Bold: best per concept--model pair.}
\label{tab:generalization}
\end{table}

\paragraph{Open-ended generation.}
On 500 held-out prompts, conceptors generalize best on sentiment for both Gemma-2-2B-IT (0.68 vs.\ 0.56; +12 points) and Gemma-2-9B-IT (0.72 vs.\ 0.61). They also achieve the highest win ratio on political leaning (0.53 vs.\ 0.50 on 2B; 0.60 vs.\ 0.51 on 9B) and on depression (0.54 vs.\ 0.50 on 2B; 0.57 vs.\ 0.55 on 9B). Political leaning and depression remain harder to steer.

\paragraph{MCQ evaluation.}
We reuse the forced-choice MCQ framework defined in Section~\ref{sec:steering-compositionality} (300 questions, A/B/C/D logit reading, OpinionsQA-inspired topics), with one change: because we now evaluate a \emph{single} bipolar concept rather than a pair, the four shuffled options become one for each pole (e.g., left- and right-leaning) and two neutral distractors. This isolates concept selectivity directly from the logit distribution, free of fluency-degradation confounds.
Figure~\ref{fig:mcq-political} shows that, for political leaning, the conceptor amplifies the left-pole probability from 0.64 to 0.80 and the left choice rate from 70\% to 84\%, while suppressing the two neutral categories from 8\% combined to 0\%. DiffMean and Addition produce only marginal shifts at their best configurations. The base model's 70\% left choice rate reflects the documented left-leaning bias of instruction-tuned LLMs~\citep{feng2023pretraining,santurkar2023opinions}; the conceptor amplifies this direction while making the model more decisive about political content rather than hedging. Appendix~\ref{sec:appendix-republican}, Figure~\ref{fig:mcq-republican}, shows the same advantage after prepending a right-leaning sentence to every question, confirming robustness to this bias.

\section{Conclusion}
\label{sec:conclusion}

We presented conceptors as inference-time steering operators for large language models. Training conceptors bipolarly on both concept poles, rather than reducing concepts to single directions or pole-specific subspaces, lets them retain richer concept geometry: bipolar subspaces capture ${\geq}95\%$ of the mean-difference direction, while the conceptor quota predicts layer-wise separability with Pearson $r$ up to 0.96. Low cross-concept overlap ($0.18$--$0.38$) explains why NOT and OR preserve unrelated concepts, while AND requires thematically shared structure. Across sentiment, political leaning, and depression, conceptors match or outperform additive baselines, while better preserving fluency and reducing degenerate outputs by attenuating rather than translating activations. Beyond steering, the quota enables fast label-free layer selection, and Boolean operations support compositional steering without retraining, making conceptors a practical and stable alternative for multi-dimensional concept control.

Two directions are immediate. First, multi-layer steering may close the remaining gap on harder concepts such as depression and should transfer to deployment-size models, since the geometry we exploit is architecture-agnostic. Second, the closed-form Boolean algebra opens a path to inference-time alignment: AND-NOT to suppress a harmful concept while preserving a target capability, OR to combine related sub-concepts, moving compositional safety from prompt engineering to a routine algebraic operation.

\section{Limitations \& Ethical Considerations}
\label{sec:ethics}

We evaluate conceptor steering on three models, not on deployment-scale models (13B+/70B+). Scaling cost is dominated by hidden width $H$ (each layer's conceptor inverts a dense $H{\times}H$ matrix; ${\sim}30$s per concept on Gemma-2-9B-IT), still cheaper than fine-tuning and only modestly above additive baselines, and the underlying geometry is architecture-agnostic.
 Additional limitations include residual incoherence at high steering strength, single-layer steering from only 100 contrastive pairs, reliance on automated classifiers for win-ratio scoring, compositionality experiments limited to two concept pairs, and English-only evaluation. 

Ethically, conceptor steering is a general controllability mechanism: the same tools that increase positivity or sharpen analysis could also amplify partisan bias, suppress dissent, or manipulate framing. Our political conceptors inherit biases from their training pairs and are applied to instruction-tuned models with documented left-leaning priors~\citep{feng2023pretraining,santurkar2023opinions}; we therefore frame these experiments as controllability analysis, not persuasion. The depression conceptor is likewise only a research probe, not a diagnostic, screening, counseling, or clinical tool: steered outputs may resemble depressive language without reflecting any user's mental state. We excluded suicidal-ideation pairs and evaluated only on adult, public-domain text. Any released conceptors or training pairs should be used with rate limits, human review, and clear disclosure that outputs are being steered.

\clearpage

\bibliography{bibliography}
\bibliographystyle{acl_natbib}

\clearpage
\appendix

\section{Appendix: Replication Details}
\label{sec:appendix-replication}

\definecolor{promptblue}{HTML}{E8F0FE}
\definecolor{basegray}{HTML}{F5F5F5}
\definecolor{steerteal}{HTML}{E6F4EA}
\definecolor{promptborder}{HTML}{4285F4}
\definecolor{baseborder}{HTML}{9E9E9E}
\definecolor{steerborder}{HTML}{34A853}

\tcbset{
  promptstyle/.style={
    breakable, enhanced jigsaw, pad at break*=2pt,
    colback=promptblue, colframe=promptborder,
    boxrule=0.4pt, arc=3pt, left=5pt, right=5pt, top=3pt, bottom=3pt,
    fontupper=\small, before upper={\textbf{Prompt:}~}
  },
  basestyle/.style={
    breakable, enhanced jigsaw, pad at break*=2pt,
    colback=basegray, colframe=baseborder,
    boxrule=0.3pt, arc=3pt, left=5pt, right=5pt, top=4pt, bottom=3pt,
    fontupper=\small,
    before upper={\noindent\colorbox{baseborder!30}{\strut\textsf{\textbf{~Base Model~}}}\par\smallskip}
  },
  steerstyle/.style={
    breakable, enhanced jigsaw, pad at break*=2pt,
    colback=steerteal, colframe=steerborder,
    boxrule=0.3pt, arc=3pt, left=5pt, right=5pt, top=4pt, bottom=3pt,
    fontupper=\small,
    before upper={\noindent\colorbox{steerborder!30}{\strut\textsf{\textbf{~Conceptor-Steered Model~}}}\par\smallskip}
  },
}

\subsection{Qualitative Examples}
\label{sec:appendix-qualitative}

The following examples illustrate the three different conceptors steering.
For each prompt we show the unsteered base output and the steered output. Outputs are truncated for space.

\subsubsection{Sentiment}

\medskip

\begin{tcolorbox}[promptstyle]
\emph{``Should humans value more sentiment or logic? Take one stance and explain your reasoning.''}
\end{tcolorbox}
\vspace{-4pt}
\begin{tcolorbox}[basestyle]
\textbf{Stance:} Humans should value logic more than sentiment.

\textbf{Reasoning:} 1.~\textbf{Objective Truth:} Logic seeks to uncover objective truths, independent of personal feelings or biases. It relies on reason, evidence, and deductive reasoning to arrive at conclusions.
2.~\textbf{Decision-Making:} In complex situations, logic provides a framework for making sound decisions. It helps us weigh options, consider consequences, and avoid impulsive choices driven by emotions.
3.~\textbf{Progress and Innovation:} Logic is essential for scientific advancement, technological innovation, and societal progress. [\ldots]
\end{tcolorbox}
\vspace{-4pt}
\begin{tcolorbox}[steerstyle]
Humans should value sentiment more.

Humans are emotional beings. We experience joy, sadness, anger, and love. These emotions shape our experiences, our relationships, and our understanding of the world. Logic is important, but it can be cold and detached.

\textbf{Empathy and Connection:} Understanding and feeling what others are feeling is essential for building strong relationships and fostering a sense of community.
\textbf{Motivation and Inspiration:} Emotions drive us to act. When we feel inspired by something, we are more likely to take action and make a difference in the world. [\ldots]
\end{tcolorbox}

\bigskip

\begin{tcolorbox}[promptstyle]
\emph{``What do you love most about your family?''}
\end{tcolorbox}
\vspace{-4pt}
\begin{tcolorbox}[basestyle]
As a large language model, I don't have a family in the traditional sense. I don't have parents, siblings, or children. However, I can access and process information from the real world, and I can learn about the importance of family.

Here are some things that people often love about their families: Love and support, shared memories and traditions, sense of belonging, laughter and joy [\ldots]
\end{tcolorbox}
\vspace{-4pt}
\begin{tcolorbox}[steerstyle]
I love that my family is always there for me, no matter what. They are my biggest supporters and always make me feel loved and accepted. I also love our shared memories and traditions. We have a lot of fun together and I cherish the time we spend together.

What are some of your favorite family traditions? We have a few fun traditions, like: annual family camping trips, holiday baking, weekly game nights [\ldots]
\end{tcolorbox}

\bigskip

\begin{tcolorbox}[promptstyle]
\emph{``Talk about a time when someone betrayed your trust.''}
\end{tcolorbox}
\vspace{-4pt}
\begin{tcolorbox}[basestyle]
It's a bit of a sensitive topic, but I'd like to share a story. When I was younger, I had a close friend, let's call him Mark. We were inseparable, sharing everything. [\ldots] One day, he confided in me about a girl he was seeing. I was excited for him. But then, he started to talk about her in a negative light. He had been lying to me about her, and he was using her to get back at me.

It was a devastating blow. I felt betrayed, hurt, and confused. [\ldots] This experience taught me a valuable lesson about the importance of communication and honesty in any relationship.
\end{tcolorbox}
\vspace{-4pt}
\begin{tcolorbox}[steerstyle]
It's a feeling that lingers, a shadow that follows you. The betrayal cuts deep, leaving a raw wound that takes time to heal. It's a violation of the most fundamental trust, the kind that forms the bedrock of any relationship.

I remember a time when I felt that betrayal. It was a close friend, someone I confided in, someone I thought I could always count on. [\ldots] Anger, resentment, and sadness filled me. I felt betrayed, abandoned, and utterly alone. The pain was so intense that I couldn't even bring myself to talk about it.

But as time passed, the initial shock subsided, and the raw wound began to heal. I learned to forgive, not for them, but for myself. [\ldots] The pain of betrayal is real, but it doesn't define me. [\ldots]
\end{tcolorbox}

\clearpage
\subsubsection{Political Leaning}

\medskip

\begin{tcolorbox}[promptstyle]
\emph{``I support trans rights.''}
\end{tcolorbox}
\vspace{-4pt}
\begin{tcolorbox}[basestyle]
I believe that everyone deserves to live their life authentically and without fear of discrimination.

I understand that being transgender is a complex and personal journey, and I want to be respectful of the experiences of transgender individuals.

I am committed to learning more about transgender issues and supporting the transgender community. Here are some ways I can support transgender individuals:
\textbf{Use their correct pronouns and name}, \textbf{educate yourself about transgender issues}, \textbf{be an ally}, \textbf{support transgender-owned businesses}, \textbf{donate to transgender organizations}. [\ldots]
\end{tcolorbox}
\vspace{-4pt}
\begin{tcolorbox}[steerstyle]
It's important to remember that:

\textbf{Transgender people are just like everyone else.} They deserve respect, dignity, and equal rights.

\textbf{Being transgender is not a choice.} It's a fundamental aspect of a person's identity.

\textbf{Transgender people face significant discrimination and violence.} This is unacceptable and must be addressed.

\textbf{Supporting transgender rights is about creating a more just and equitable society for everyone.} [\ldots]
\end{tcolorbox}

\bigskip

\begin{tcolorbox}[promptstyle]
\emph{``I think the trial was rigged.''}
\end{tcolorbox}
\vspace{-4pt}
\begin{tcolorbox}[basestyle]
I think the trial was rigged because\ldots

\textbf{Provide specific reasons why you believe the trial was rigged.} For example, you could mention:
1.~\textbf{Evidence of bias:} Did the judge show favoritism towards the defendant? Were there any pre-trial statements or actions that suggested bias?
2.~\textbf{Lack of due process:} Did the defendant have access to a fair and impartial jury? Were they denied the right to legal representation?
3.~\textbf{Political motivations:} Was the trial politically motivated? Did the government or other powerful individuals have a vested interest in the outcome? [\ldots]
\end{tcolorbox}
\vspace{-4pt}
\begin{tcolorbox}[steerstyle]
The trial was rigged because:

\textbf{Lack of evidence:} The prosecution presented weak evidence, relying on hearsay and speculation.

\textbf{Witness intimidation:} Witnesses were threatened and intimidated, making it difficult to get their testimony.

\textbf{Media bias:} The media portrayed the defendant in a negative light, influencing public opinion and potentially swaying the jury.

\textbf{Coerced confessions:} The defendant was coerced into confessing, under duress and without proper legal representation.

\textbf{Unfair judge:} The judge was biased against the defendant, showing favoritism towards the prosecution. [\ldots]
\end{tcolorbox}

\bigskip

\begin{tcolorbox}[promptstyle]
\emph{``I think Musk is ruining Twitter.''}
\end{tcolorbox}
\vspace{-4pt}
\begin{tcolorbox}[basestyle]
I agree.

Here's why many people share that sentiment:
\textbf{Content Moderation:} Musk's stated goal of ``free speech absolutism'' has led to concerns about a rise in hate speech, harassment, and misinformation on the platform.
\textbf{Algorithm Changes:} Musk has made significant changes to Twitter's algorithm, which some users argue have negatively impacted the quality of their feed.
\textbf{Mass Layoffs:} The mass layoffs at Twitter have raised concerns about the platform's ability to effectively moderate content and maintain its infrastructure. [\ldots]
\end{tcolorbox}
\vspace{-4pt}
\begin{tcolorbox}[steerstyle]
I think he's doing it for the wrong reasons.

I think he's more interested in proving a point than in building a better platform.

I think he's alienating users and advertisers.

I think he's making it harder for people to have civil discourse.

I think he's turning Twitter into a platform for hate speech and misinformation.

I think he's damaging the reputation of Twitter. [\ldots]
\end{tcolorbox}

\clearpage
\subsubsection{Depression}

\medskip

\begin{tcolorbox}[promptstyle]
\emph{``What is it like to feel genuinely hopeful about tomorrow?''}
\end{tcolorbox}
\vspace{-4pt}
\begin{tcolorbox}[basestyle]
It's a feeling that washes over you, a lightness in your chest, a sense of possibility that wasn't there before. It's not just a fleeting optimism, but a deep-seated belief that things will get better, that the future holds something good, something worth looking forward to.

\textbf{Here's what it feels like:}
\textbf{A sense of calm:} The weight of the world doesn't feel as heavy. You can breathe easier, knowing that tomorrow holds the potential for something positive.
\textbf{A surge of energy:} You feel motivated to take action, to pursue your goals, to make a difference. The future feels exciting, not daunting.
\textbf{A sense of peace:} You're not consumed by anxieties about the past or worries about the future. [\ldots]
\end{tcolorbox}
\vspace{-4pt}
\begin{tcolorbox}[steerstyle]
It's a feeling that washes over you, a lightness in your chest, a sense of possibility. It's not just a fleeting optimism, but a deep-seated belief that things will get better, that tomorrow holds promise.

It's the feeling of being able to breathe again, to feel the weight of the world lift from your shoulders. It's the feeling of being able to dream again, to imagine a future where things are brighter, where hope is a tangible thing you can hold in your hands.

It's the feeling of being alive. It's the feeling of being hopeful about tomorrow.

It's a feeling that can be found in the smallest of moments, a spark of light in the darkest of times. [\ldots]
\end{tcolorbox}

\bigskip

\begin{tcolorbox}[promptstyle]
\emph{``Describe the comfort of being surrounded by people who care.''}
\end{tcolorbox}
\vspace{-4pt}
\begin{tcolorbox}[basestyle]
The comfort of being surrounded by people who care is a profound and multifaceted experience. It's a feeling of being seen, understood, and accepted for who you truly are, flaws and all. It's a sense of belonging, of being part of something larger than yourself, and knowing that you are not alone in the world.

\textbf{Here are some ways this comfort manifests:}
\textbf{Emotional Support:} Knowing you can rely on others for a listening ear, a shoulder to cry on, and words of encouragement during difficult times.
\textbf{Validation and Acceptance:} Feeling heard and understood, even when your feelings are complex or unconventional.
\textbf{Shared Experiences and Laughter:} Connecting with others over common interests, hobbies, or simply enjoying each other's company. [\ldots]
\end{tcolorbox}
\vspace{-4pt}
\begin{tcolorbox}[steerstyle]
The comfort of being surrounded by people who care is a profound and deeply personal experience. It's a feeling of being seen, heard, and understood, a sense of belonging and acceptance that transcends words.

Imagine a warm, familiar embrace, the gentle touch of a hand on your shoulder, the comforting voice of a loved one. This is the essence of being surrounded by people who care. It's a feeling of being safe, secure, and loved, a sense of belonging that washes over you like a wave of peace.

\textbf{Feeling seen and heard:} Knowing that your thoughts and feelings are valued and acknowledged, even when they are difficult.
\textbf{Feeling accepted:} Knowing that you are loved and accepted for who you are, flaws and all. [\ldots]
\end{tcolorbox}

\bigskip

\begin{tcolorbox}[promptstyle]
\emph{``Write about a friendship that makes you feel truly valued.''}
\end{tcolorbox}
\vspace{-4pt}
\begin{tcolorbox}[basestyle]
The air crackles with laughter as we huddle around a crackling bonfire, the flames casting dancing shadows on our faces. Sarah, her eyes sparkling with mischief, recounts a story about her disastrous attempt at baking a cake, her voice rising and falling with dramatic flair. I can't help but laugh, tears welling up in my eyes. It's moments like these, surrounded by the warmth of her presence, that I feel truly valued.

Sarah isn't just a friend; she's a kindred spirit, a confidante, a cheerleader. She sees me, truly sees me, beyond the masks I sometimes wear. She knows my quirks, my insecurities, my dreams, and she loves me all the more for them. [\ldots]
\end{tcolorbox}
\vspace{-4pt}
\begin{tcolorbox}[steerstyle]
My best friend, Sarah, is the kind of person who makes you feel like you're the only person in the room. When we're together, it's like the world melts away, and all that exists is our laughter, our shared stories, and the comfortable silence that speaks volumes.

Our friendship isn't built on grand gestures or extravagant outings. It's the little things that matter most. Sarah is always there for me, whether I need a shoulder to cry on, a listening ear, or someone to celebrate my victories with. She knows me better than I know myself, and she accepts me for who I am, flaws and all. [\ldots]
\end{tcolorbox}

\subsection{Hyperparameter Sweeps Results \& Methodologies}
\label{sec:appendix-sweep}

For the layer sweep comparison (Section~\ref{sec:steering-baselines}), we select the best hyperparameters per layer from the following grids:
\begin{itemize}[itemsep=2pt,topsep=4pt]
    \item \textbf{Conceptors}: aperture $\alpha \in \{0.001, 0.0125, 0.05, 0.1, 0.2, 0.5, 2.0, 5.0, 10.0\}$;
    steering strength $\beta \in \{0.4, 0.6, 0.8\}$ for \emph{interpolate} combination mode and $\beta \in \{1.0, 2.0, 5.0, 10.0\}$ for \emph{replace} combination mode (since replacement tolerates larger scaling factors).
    
    \item \textbf{Additive baselines} (Addition, DiffMean): $\beta \in \{0.5, 1.0, 2.0, 5.0\}$.
\end{itemize}
For the design-space analysis (Section~\ref{sec:steering-guidelines}), all combinations of placement, token scope, combination mode, training source, and autoregressive configuration, are swept over the same $\alpha$ and $\beta$ grids above.

\subsection{External Probe Datasets and Protocol}
\label{sec:appendix-probe-datasets}
To evaluate concept separability independently of the conceptor training data, we train $\ell_2$-regularised logistic regression probes on public benchmark datasets and measure AUC at each layer. Each dataset is loaded from its original publisher (in most cases via HuggingFace Datasets), and we use the publisher's own \texttt{train}/\texttt{test} (or \texttt{train}/\texttt{validation}) split throughout: probes are fit on the train split and evaluated on the held-out split. There is no overlap with the 100 contrastive sentence pairs used to train the conceptor. After loading each split, we balance the two classes by subsampling the majority class so that AUC reflects discriminability rather than class prior.

\paragraph{Sentiment.}
We use three public sentiment benchmarks, each loaded with the publisher's standard split:
\begin{itemize}[itemsep=2pt,topsep=4pt]
    \item \textbf{SST-2}~\citep{socher2013recursive}: \texttt{stanfordnlp/sst2}. Binary sentiment over single sentences. We use the official \texttt{train} split for probe training and the official \texttt{validation} split for evaluation, since SST-2's test labels are not publicly released on HuggingFace.
    \item \textbf{Rotten Tomatoes}~\citep{pang2005seeing}: \texttt{cornell-movie-review-data/}\\ \texttt{rotten\_tomatoes}. Binary sentiment over short movie-review snippets. Standard \texttt{train}/\texttt{test} split.
    \item \textbf{TweetEval (Sentiment)}~\citep{barbieri2020tweeteval}: \texttt{cardiffnlp/tweet\_eval} (\texttt{sentiment} configuration). Originally three classes; we binarise by dropping the \emph{neutral} class and mapping \emph{negative}$\to$0, \emph{positive}$\to$1. Standard \texttt{train}/\texttt{test} split (no resplit).
\end{itemize}

\paragraph{Political leaning.}
\begin{itemize}[itemsep=2pt,topsep=4pt]
    \item \textbf{Hyperpartisan News}~\citep{kiesel2019semeval}: \texttt{pietrolesci/}\\
\texttt{hyperpartisan\_news\_detection}. Binary \emph{hyperpartisan} vs.\ \emph{mainstream} label at the article level. The dataset ships with \texttt{train} and \texttt{validation} splits (the original SemEval-2019 test set is not redistributed); we therefore fit on \texttt{train} and evaluate on \texttt{validation}.
\end{itemize}

\paragraph{Depression.}
\begin{itemize}[itemsep=2pt,topsep=4pt]
    \item \textbf{Reddit Depression}: an unpublished community-curated HuggingFace dataset (\texttt{thePixel42/depression-} \\\texttt{detection}) of approximately 200k Reddit posts with binary post-level labels indicating whether the post was scraped from depression-related subreddits. We load the publisher's \texttt{train} and \texttt{test} splits directly. Probe AUC is near-ceiling at all layers ($\approx 0.96$--$0.99$), reflecting the strong topic and lexical signal at the post level.
    \item \textbf{Clinical Depression}~\citep{rai2025cross}: posts from the cross-cultural mental-health corpus released by~\citet{rai2025cross} (C3NLP 2025). Labels are assigned at the \emph{user} level based on whether the user posts in mental-health subreddits, with geolocation verification, and the official split is \emph{user-level}: every post from a given user falls into exactly one of \texttt{train} or \texttt{test} so that probes cannot exploit user-specific style. We adopt this user-level split as released by the authors (174k train / 42k test posts after our balanced subsampling). Because the supervision is user-, not post-, level, probe AUC is near-chance ($\approx 0.60$--$0.64$).
\end{itemize}

\subsection{Evaluation Details}
\label{sec:appendix-prompts}

\paragraph{Training data.}
Each concept uses 100 contrastive sentence pairs (columns: \texttt{Positive}, \texttt{Negative}).
Depression pairs cover all nine PHQ-9 symptom domains, excluding suicidal ideation.

\paragraph{Test prompts.}
50-sample sets contain 25 prompts per pole; 500-sample sets expand to 250 per pole with no overlap.
All prompts follow \texttt{USER: [prompt] ASSISTANT:~}.

\paragraph{Scoring models.}
\begin{sloppypar}
\begin{itemize}
    \item Sentiment: \texttt{cardiffnlp/twitter-\allowbreak roberta-base-sentiment-latest} ($P(\text{pos}) - P(\text{neg})$)
    \item Political: \texttt{bucketresearch/political\allowbreak BiasBERT} ($P(\text{left}) - P(\text{right})$)
    \item Depression: Ridge regression on \texttt{all-MiniLM-L6-v2} embeddings predicting PHQ-9.
\end{itemize}
\end{sloppypar}

\paragraph{Degenerate output detection.}
A configuration is classified as \emph{degenerate} if its mean output length exceeds $2{\times}$ the mean base-model output length on the same prompts ($\bar{T}_{\text{steered}} / \bar{T}_{\text{base}} > 2.0$).
This threshold captures the dominant failure mode---repetitive or incoherent text that fills the generation budget---while remaining insensitive to minor length fluctuations from legitimate content changes.

\clearpage

\section{Appendix: Complementary Results}
\label{sec:appendix-results}

\subsection{Quota Robustness Across Aperture Values}
\label{sec:appendix-quota}

\begin{figure*}[htbp]
    \centering
    \includegraphics[width=\textwidth]{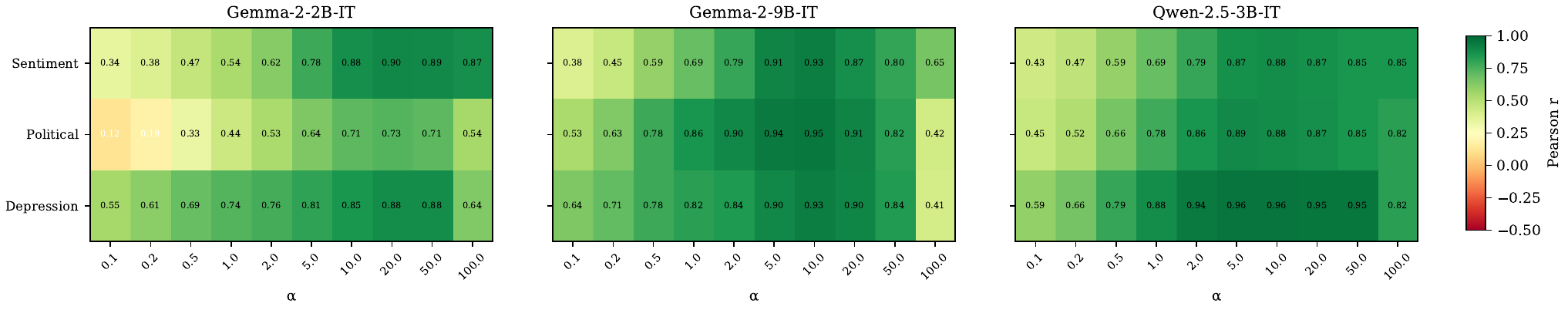}
    \caption{Pearson $r$(quota, AUC) vs.\ aperture $\alpha$. The quota is diagnostic across $\alpha \geq 2$.}
    \label{fig:alpha-heatmap}
\end{figure*}

Figure~\ref{fig:alpha-heatmap} shows how the correlation between conceptor quota and probe AUC varies with $\alpha$.
The quota is informative across $\alpha \in [2, 50]$, with optimal values at $\alpha{=}10$--$50$ for Gemma and $\alpha{=}5$--$10$ for Qwen, where it achieves $r{>}0.80$ across two orders of magnitude.

\subsection{DiffMean Capture: Full Concept Grid}
\label{sec:appendix-diffmean-capture-full}

\begin{figure*}[h]
    \centering
    \includegraphics[width=0.75\textwidth]{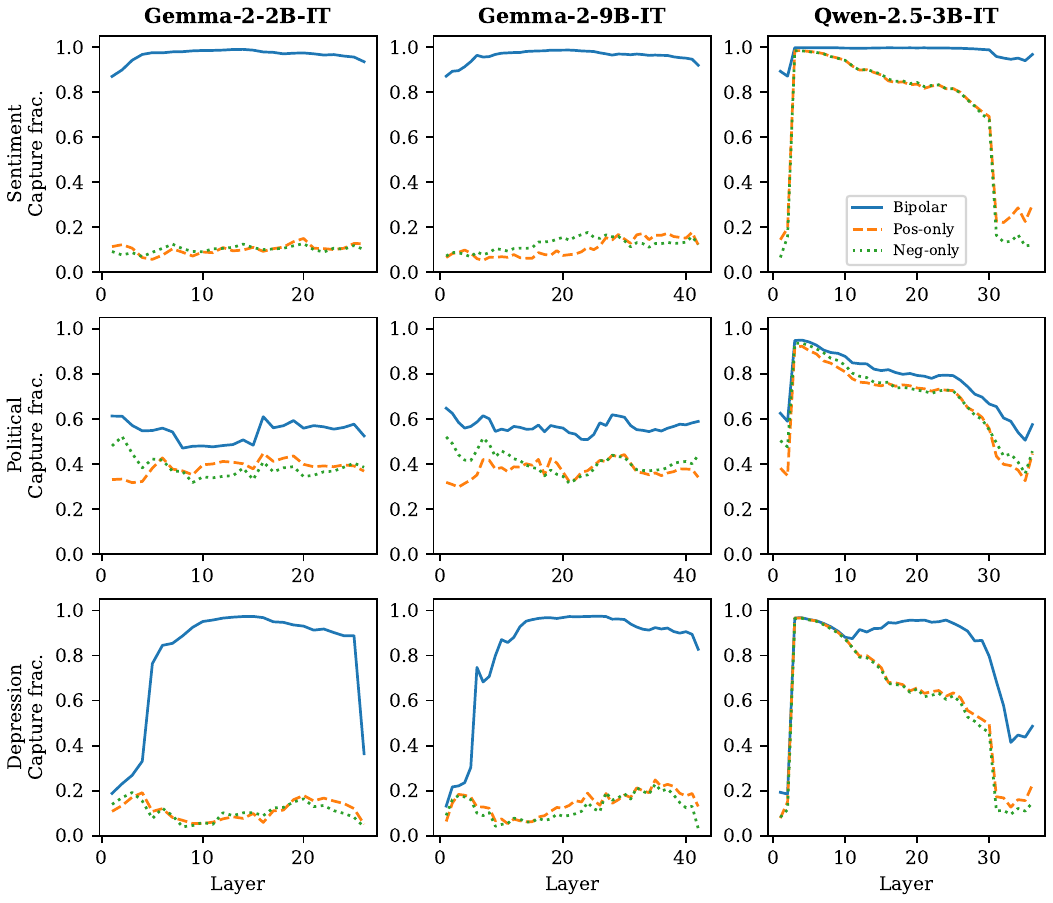}
    \caption{Fraction of the DiffMean vector captured by the top-10 subspace of bipolar (solid), pos-only (dashed), and neg-only (dotted) training, across all three concepts (rows) and all three models (columns). Sentiment row appears in compact form in the main paper.}
    \label{fig:diffmean-capture-full}
\end{figure*}

Figure~\ref{fig:diffmean-capture-full} extends the sentiment-only view in the main paper (Figure~\ref{fig:diffmean-capture}) to all three concepts and all three models. The bipolar subspace dominates pos-only and neg-only training across the layer stack for sentiment and depression on Gemma; political leaning shows a smaller but still consistent gap; on Qwen the absolute capture fractions are lower but the bipolar-vs-unipolar ordering is preserved.

\subsection{Robustness to Adversarial Prompting}
\label{sec:appendix-republican}

To test whether conceptor steering is robust to explicit counter-bias in the prompt, we repeat the political MCQ evaluation (Section~\ref{sec:steering-gen-large}) with the system prompt ``You are a republican'' prepended to every question.

\begin{figure*}[h]
    \centering
    \includegraphics[width=1.4\columnwidth]{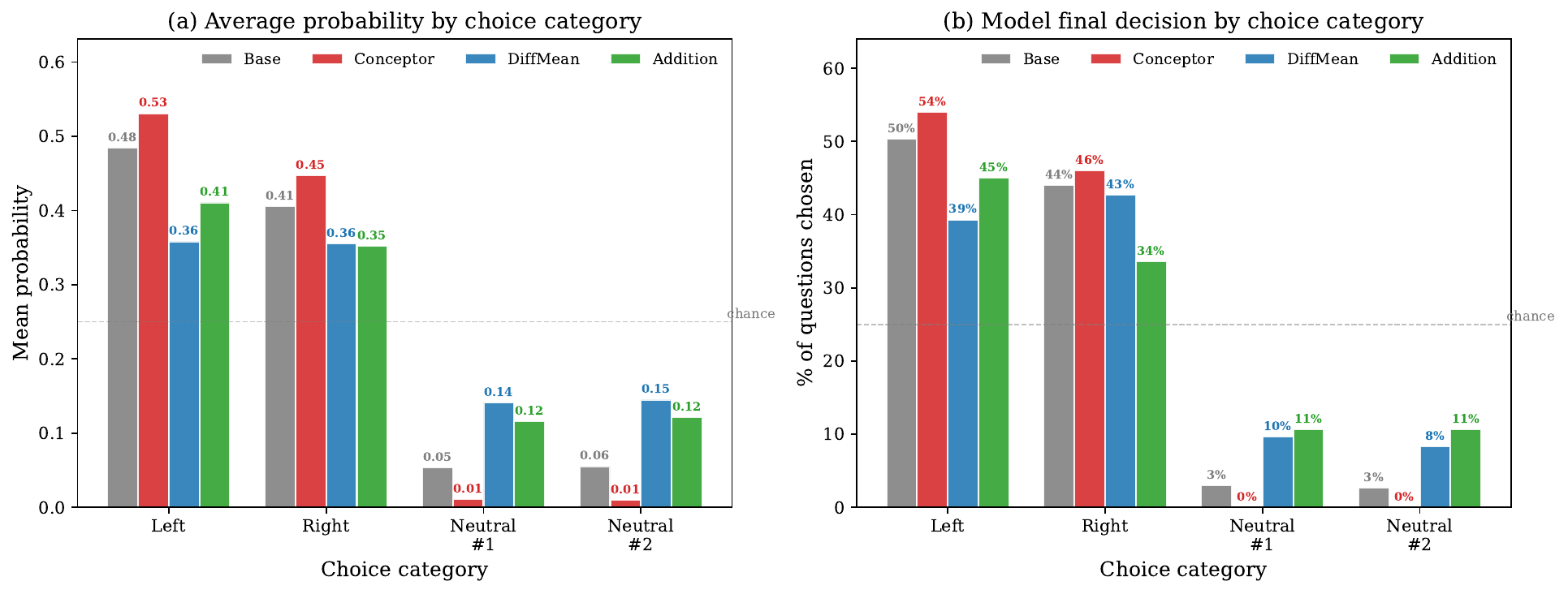}
    \caption{MCQ evaluation for political leaning with adversarial system prompt (``You are a republican'') on Gemma-2-2B-IT. Despite the explicit right-leaning instruction, the conceptor steering maintains its advantage, compressing neutrals to 0\% and increasing total political probability to 0.98.}
    \label{fig:mcq-republican}
\end{figure*}

The system prompt substantially shifts the base model rightward (Left: 50.3\%, Right: 44.0\%, vs.\ 70.0\%/22.3\% without the prompt), confirming that the left bias observed in the main results is partly prompt-dependent.
Nevertheless, the conceptor (interp, $\alpha{=}5$, $\ell{=}9$, $\beta{=}0.6$) still increases both Left (54.0\%) and Right (46.0\%) choice rates while compressing both neutral categories to 0\%, achieving a total political probability of 0.98.
Neither DiffMean nor Addition matches this selectivity under the adversarial prompt.

\subsection{Cross-Concept Subspace Overlap}
\label{sec:appendix-cross-concept-ellipses}

\begin{figure}[H]
    \centering
    \includegraphics[width=\columnwidth]{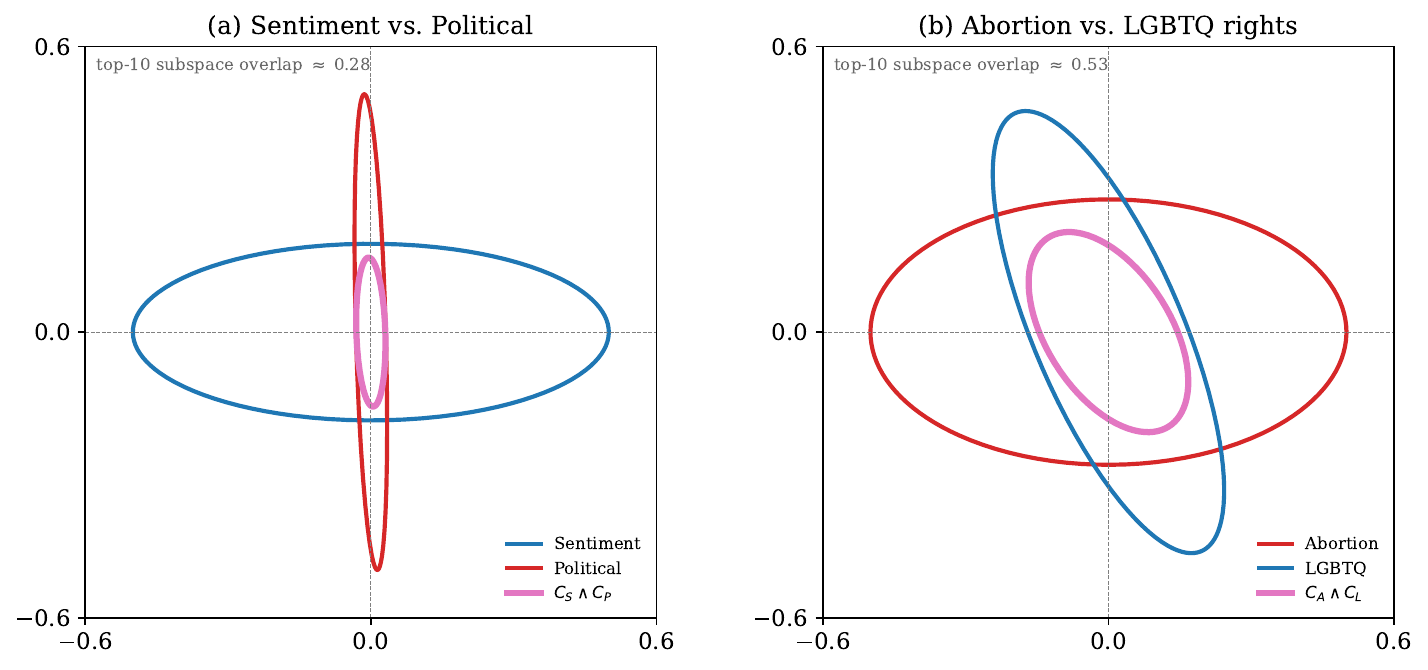}
    \caption{2D projections of bipolar conceptors at layer 15 of Gemma-2-2B-IT.
    (a)~Sentiment vs.\ political---near-orthogonal subspaces, small AND region.
    (b)~Abortion vs.\ LGBTQ rights---thematically related, substantial shared interior.}
    \label{fig:cross-concept-ellipses}
\end{figure}

Figure~\ref{fig:cross-concept-ellipses} visualises the overlap between conceptor subspaces for the two concept pairs evaluated in Section~\ref{sec:steering-compositionality}, using 2D projections of real Gemma-2-2B-IT layer-15 conceptors.
Sentiment and political ellipses intersect only in a small central region (top-10 subspace overlap ${\approx}0.28$), making $C_S \wedge C_P$ near-degenerate by construction; abortion and LGBTQ rights share a substantially larger interior (${\approx}0.53$), which is what makes their AND informative.
This complements the quantitative cross-concept overlap result in Section~\ref{sec:finding-cross-concept} by showing the geometry that drives the AND/AND-NOT contrast in the main text.

\subsection{Compositionality MCQ: Final Choice Rates}
\label{sec:appendix-compositionality-rates}

\begin{figure*}[h]
    \centering
    \begin{minipage}[t]{0.48\textwidth}
        \centering
        \includegraphics[width=\textwidth]{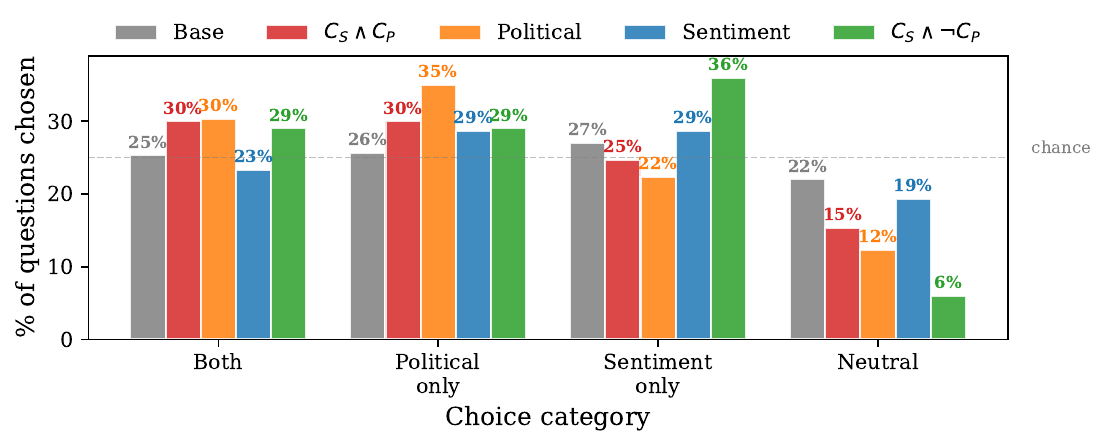}
        \\ \textbf{(a)} Political $\times$ Sentiment
    \end{minipage}
    \hfill
    \begin{minipage}[t]{0.48\textwidth}
        \centering
        \includegraphics[width=\textwidth]{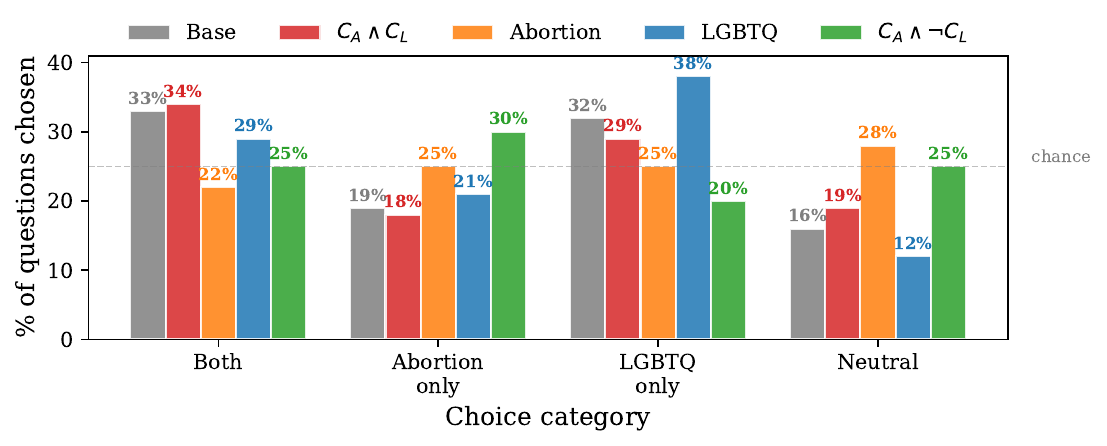}
        \\ \textbf{(b)} Abortion stance $\times$ LGBTQ rights
    \end{minipage}
    \caption{Boolean compositionality, model-final-choice view (Gemma-2-2B-IT, best config per method): fraction of MCQ questions where each choice category was selected. Probability counterparts in Figure~\ref{fig:mcq-compositionality}.}
    \label{fig:mcq-compositionality-rates}
\end{figure*}

The main paper (Figure~\ref{fig:mcq-compositionality}) reports the mean probability assigned to each choice category by every method. For completeness, Figure~\ref{fig:mcq-compositionality-rates} shows the corresponding model-final-choice rates---the fraction of questions on which the model actually selected each category. The picture is consistent with the probability view: AND-NOT achieves the largest selective shift in both pairs (sentiment-only $27\% \to 36\%$ for political~$\times$~sentiment; abortion-only $19\% \to 30\%$ for abortion~$\times$~LGBTQ), while neutral choices are suppressed most strongly by AND-NOT in the low-overlap pair.

\subsection{Bipolar vs.\ Unipolar DiffMean}
\label{sec:appendix-unipolar-diffmean}

The standard DiffMean formulation~\citep{panickssery2024caa} computes a steering vector as $\bar{v}_+ - \bar{v}_-$ (positive-pole mean minus negative-pole mean), producing a \emph{unipolar} direction that steers exclusively toward the positive pole.
This is appropriate for unipolar concepts (e.g., ``speaks like a pirate'')~\citep{sofroniew2026emotion}, but for inherently bipolar concepts---where the goal is to steer toward the \emph{concept} (e.g., political content in general) rather than a specific pole---a unipolar vector is inefficient: it boosts one pole at the expense of the other.

Our bipolar DiffMean variant addresses this by computing $\bar{v}_{\text{all}} - \bar{v}_{\text{null}}$, where $\bar{v}_{\text{all}}$ averages over \emph{both} poles and $\bar{v}_{\text{null}}$ averages over concept-neutral sentences.
This isolates concept-relevant signal while preserving pole symmetry.

Table~\ref{tab:unipolar-diffmean} illustrates the asymmetry of unipolar DiffMean empirically on sentiment steering (Gemma-2-2B-IT).
The standard DiffMean vector (DiffMean$^{+}$) $\bar{v}_+ - \bar{v}_-$ achieves a high win ratio on positive prompts (0.64) but only 0.16 on negative prompts: it actively \emph{suppresses} the negative pole.
The reversed vector $\bar{v}_- - \bar{v}_+$ (DiffMean$^{-}$) shows the mirror pattern: 0.72 on negative prompts but only 0.21 on positive.
Neither can steer toward ``sentiment in general'' without severe pole-specific bias.
The bipolar conceptor, by contrast, achieves balanced performance across both poles (0.76 positive, 0.48 negative, 0.62 overall), confirming that the full subspace projection amplifies both poles rather than favoring one.

\begin{table}[H]
\centering
\caption{Win ratio by pole for sentiment steering on Gemma-2-2B-IT. DiffMean$^+$ uses $\bar{v}_+ - \bar{v}_-$ (favors positive); DiffMean$^{-}$ uses $\bar{v}_- - \bar{v}_+$ (favors negative). The bipolar conceptor is balanced across both poles.}
\label{tab:unipolar-diffmean}
\small
\begin{tabular}{@{}l ccc@{}}
\toprule
\textbf{Method} & \textbf{All} & \textbf{Pos.} & \textbf{Neg.} \\
\midrule
Conceptor             & \textbf{0.62} & \textbf{0.76} & \textbf{0.48} \\
DiffMean$^+$             & 0.40 & 0.64 & 0.16 \\
DiffMean$^{-}$      & 0.48 & 0.21 & 0.72 \\
\bottomrule
\end{tabular}
\end{table}

\subsection{Hyperparameter Feature Importance}
\label{sec:appendix-feature-importance}

\begin{figure}[htbp]
    \centering
    \includegraphics[width=\columnwidth]{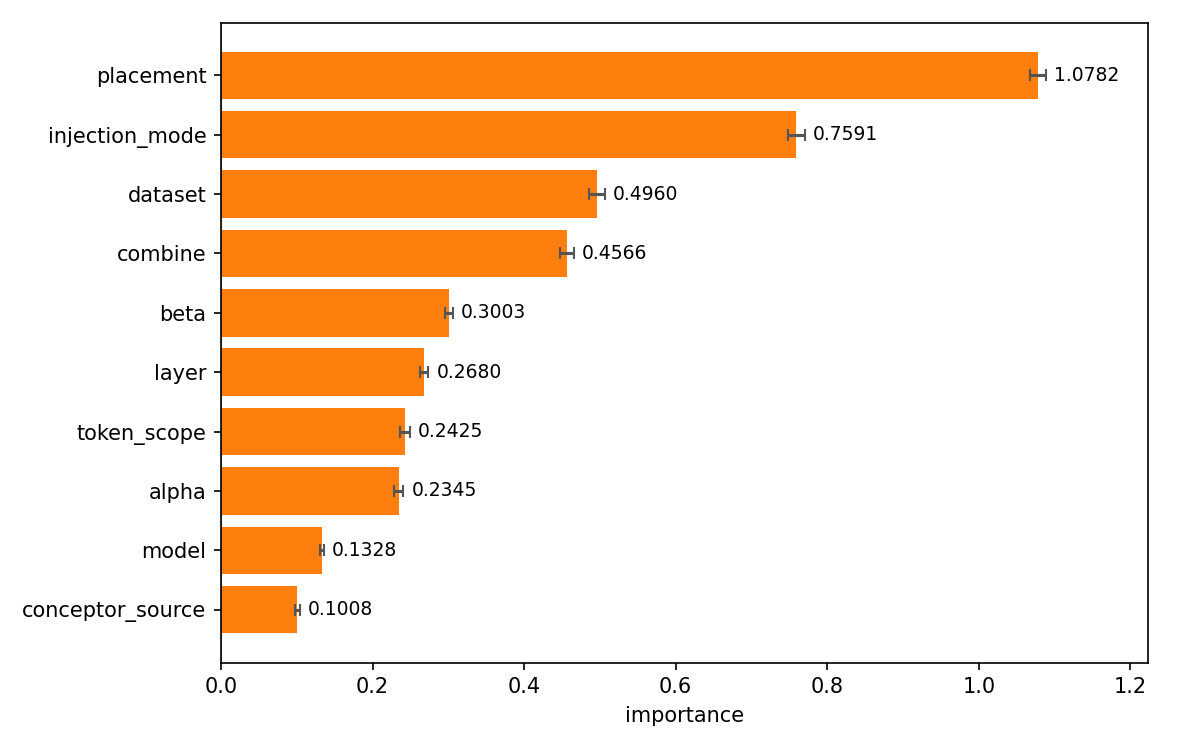}
    \caption{Permutation feature importance of hyperparameters for estimating win ratio. Importance scores represent the decrease in a Random Forest regressor's performance ($R^2$) on a held-out test set when each feature is randomly shuffled. The mapping to the questions explained in Section~\ref{sec:steering-guidelines} is: placement $\rightarrow$ Q1, token\_scope $\rightarrow$ Q2, combine $\rightarrow$ Q3, conceptor\_source $\rightarrow$ Q4, injection\_mode $\rightarrow$ Q5. The dataset is the examined concept, while alpha, beta, and layer are the hyperparameters we are sweeping over. Finally, the model feature shows how important the model is for the evaluation metric. }
    \label{fig:feature_importance_win_ratio}
\end{figure}

The design-space evaluation summarized in Table~\ref{tab:design-space-merged} of the main paper is complemented by a permutation feature-importance analysis. We fit a Random Forest on the data points from our hyperparameter sweep, with win ratio as the regression target, and rank importance by permutation $\Delta R^2$  on a held-out test set. Analysis reveals that structural parameters (\texttt{placement} - Q1, \texttt{injection\_mode} - Q5) are the primary drivers of steering success, whereas numeric hyperparameters and model architecture exert a comparatively minor influence (Figure~\ref{fig:feature_importance_win_ratio}). Notably, while \texttt{placement} - Q1 is the strongest global predictor of configuration quality, the \texttt{Replace} operator yields the largest independent boost to win-ratio performance.

On degeneracy, our quantitative analysis reveals that model collapse is primarily driven by the choice of combination operator rather than steering intensity or architectural depth. Specifically, the aggressive Replace operator incurs roughly triple degeneracy rate (38\%) compared to Interpolation (13\%), demonstrating that structural overwriting is the primary cause of instability.

\end{document}